\documentclass[11pt]{article}

\usepackage[final]{acl}

\usepackage{times}
\usepackage{latexsym}

\usepackage[T1]{fontenc}
\usepackage[utf8]{inputenc}

\usepackage{microtype}

\usepackage{inconsolata}

\usepackage{graphicx}

\usepackage{makecell} % Required for \makecell command
\usepackage{multirow}
\usepackage{tabularx}
\usepackage{caption}       
\usepackage{amsmath}
\usepackage{array}
\usepackage[table]{xcolor}
\usepackage{float} 
\usepackage{cuted}

\usepackage{booktabs}

\usepackage[T1]{fontenc} % a prerequisite that makes the [htt] option of hyphenat work reliably with monospaced fonts.
\usepackage{hyphenat} % allows breaking in \texttt

\usepackage{needspace}           % avoid orphaned section headers
\usepackage{dblfloatfix} 
\usepackage{placeins}

\usepackage[skins,breakable]{tcolorbox}
\tcbset{
  enhanced,
  breakable,
  colback=gray!5,
  colframe=black!50,
  boxrule=0.5pt,
  arc=2pt,
  outer arc=2pt,
  left=6pt,
  right=6pt,
  top=6pt,
  bottom=6pt
}

\newcommand{\green}[1]{\textcolor{green!50!black}{#1}}
\newcommand{\red}[1]{\textcolor{red!70!black}{#1}}

\title{PTEB: Towards Robust Text Embedding Evaluation via Stochastic Paraphrasing at Evaluation Time with LLMs}

\author{
    Manuel Frank \\
    Department of Computer Science \\
    Munster Technological University \\
    \texttt{Manuel.Frank@zohomail.eu} \\\And
    Haithem Afli \\
    Department of Computer Science \\
    Munster Technological University \\
    \texttt{Haithem.Afli@mtu.ie} \\
    }

\begin{document}
\maketitle
\begin{abstract}
Current sentence embedding evaluations typically rely on static test beds like the Massive Text Embedding Benchmark (MTEB). While invaluable, repeated tuning on a fixed suite can inflate reported scores and obscure real-world robustness. We introduce the Paraphrasing Text Embedding Benchmark (PTEB), a dynamic protocol that stochastically generates meaning-preserving paraphrases at evaluation time and aggregates results across multiple runs. Using a cost-efficient LLM-based method grounded in gold ratings and human validation, we show that LLMs generate token-diverse but semantically preserving paraphrases. Across 7 MTEB tasks, we validate our hypothesis that the performance of sentence encoders is sensitive to changes in token space even when semantics remain fixed. We also observe that smaller models are not disproportionately affected relative to larger ones. Our results are statistically robust over multiple runs spanning 20 datasets and 25 languages. More generally, we aim to propose a new evaluation paradigm in NLP that relies less on static, pre-defined benchmarks but shifts towards dynamic, stochastic evaluation leveraging eval-time compute. We make the code to run PTEB publicly available.\footnote{\url{https://github.com/ThisIsManuel/pteb}}
\end{abstract}

%
% ##########################################################################################
% ############################ INTRODUCTION ################################################
% ##########################################################################################

% --------------
\section{Introduction}
Text embeddings have become a cornerstone of Natural Language Processing (NLP), enabling a wide range of downstream tasks. Traditionally, embeddings are evaluated on static benchmarks like the Massive Text Embedding Benchmark (MTEB)~\citep{Muennighoff2023}, a de facto standard. 
% 
%
%  ------------------------------------------
%  ---------- FIGURE APPROACH ---------------
% -------------------------------------------
\begin{figure}[!ht]
\begin{center}
  \includegraphics[width=\columnwidth]{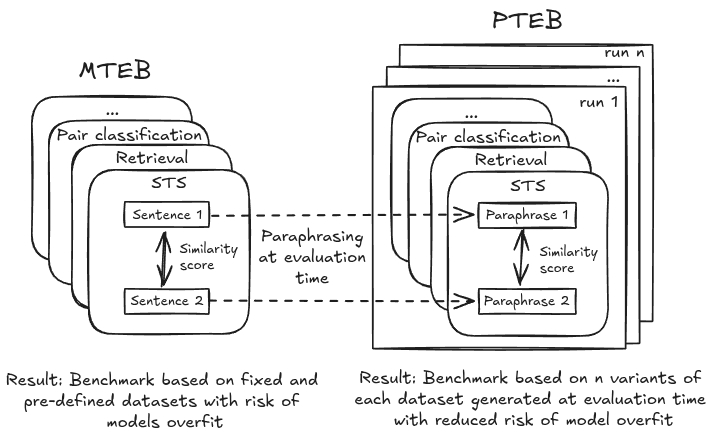}
  \caption{PTEB paraphrases data at evaluation time to measure embedding model performance.}
  \label{fig:approach}
\end{center}
\end{figure}
More recently, MTEB has been extended to the Massive Multilingual Text Embedding Benchmark (MMTEB)~\citep{MMTEB}, covering more than 250 languages, and RTEB\footnote{\url{https://huggingface.co/blog/rteb}}, a retrieval benchmark with private test sets. However, benchmarks like MTEB need to be maintained and further developed~\citep{Chung2025}. 

A persistent challenge is that static benchmarks saturate over time. Although this saturation partly reflects genuine progress, it can also be attributed to overfitting. This arises from at least two factors \citep{Liang2025}: 1. \textit{Data contamination} which refers to the inclusion of benchmark items within training data, enabling models to exploit prior exposure rather than demonstrate authentic modelling capabilities (also known as data leakage). 2. \textit{Biased overtraining} which involves deliberate allocation of training resources to the domains of anticipated benchmarks, resulting in models with uneven performance that excel in benchmarked domains while underperforming elsewhere. Overall, more robust embedding benchmarks are needed~\citep{Goel2025}. So far, the only solutions have been temporary through private test sets or the release of new or updated benchmarks.

To address this issue, we argue that model capabilities should be evaluated on dynamically generated variations of evaluation tasks at evaluation time rather than predefined static benchmarks. We therefore introduce the Paraphrasing Text Embedding Benchmark (PTEB), which generates paraphrases at eval time, creating a set of semantically equivalent but textually distinct problem instances that better approximates diverse real-world applications. While MTEB targets maximal language and task coverage, PTEB is a complementary, statistically robust protocol that stress-tests semantic invariance via multi-run eval-time paraphrasing.
 
Generative LLMs have been shown to be effective paraphrasers~\citep{Wahle2022}, e.g., for data augmentation~\citep{Dai2023}. \citet{Frank2025_GASE} and \citet{Thirukovalluru2025} used LLMs to generate variations of the input text, e.g., by paraphrasing, for data augmentation at runtime to improve the performance of embeddings. 

Prior research on robustness and benchmarking has mainly relied on algorithmic transformations, masked language modelling (MLM), or classical word embedding techniques. For example, \texttt{nlpaug}~\citep{Ma2019} provides a general-purpose augmentation toolkit with methods such as backtranslation, synonym replacement, and embedding-based substitutions (e.g., using Word2Vec~\citep{Mikolov2013}), while TextAttack~\citep{Morris2020} systematises adversarial attacks through synonym replacments, character edits, and MLM substitutions, supporting both benchmarking and adversarial training. Building on these ideas, TextFlint~\citep{Wang2021} introduced a multilingual robustness framework combining universal and task-specific transformations including adversarial attacks, all validated through human evaluation. 

More recently, \citet{Yang2023} published an article on benchmark contamination and evasion of contamination detection methods. They demonstrated the fragility of existing evaluations for generative models, showing that rephrased benchmark samples can bypass standard decontamination methods and artificially inflate the performance of a 13B parameter model to GPT-4 levels, thus undermining benchmark trustworthiness. Whereas \citet{Yang2023} use paraphrasing to diagnose weaknesses of static benchmarks for generative LLMs, we employ eval-time paraphrasing to construct contamination-resistant evaluations for embedding models. \citet{Li2025_SentenceSmith} address similar concerns using semantic graph manipulation to generate hard negatives by altering semantics, though limited to English. Existing work has therefore emphasised robustness testing, augmentation, contamination detection or semantic manipulation, but remains fundamentally constrained by static transformation rules, MLM, classical embeddings, and a focus on generative models or English corpora. 

Concurrent work by \citet{Goel2025} proposes SAGE to analyse robustness (among other dimensions) through rules-based transformations such as synonym replacement evaluated on 3 English datasets. Our approach complements this by testing robustness to richer LLM-generated paraphrases on 20 datasets with 25 languages across six random seeds validated by LLM-judges and humans. 

Our contribution is to employ generative LLMs as paraphrasers at eval-time, yielding a stochastic, dynamically moving benchmark that continuously produces novel test variants. This extends robustness testing from predefined perturbations to rich, stochastic evaluation runs, specifically targeting embedding models across diverse NLP tasks. More broadly, our goal is to shift the focus of the NLP community from static datasets and offline augmentation to stochastic evaluation-time benchmarks powered by modern generative models.  

The PTEB architecture is visualised in \autoref{fig:approach}. We make the code to run PTEB publicly available.\footnote{\url{https://github.com/ThisIsManuel/pteb}}
%
%  ------------------------------------------------------------------
%  -------------- TABLE WORD COUNTS PER DATASET ........-------------
%  ------------------------------------------------------------------
%
\begin{table}[!ht]
\small
\begin{center}
\begin{tabularx}{\columnwidth}{lX}
\toprule
\textbf{Dataset} & \textbf{Word count} \\
\midrule
BIOSSES \citep{BIOSSES}             & 22.8 \\
SICK-R \citep{SICK}                 & 9.6 \\
STS12 \citep{SemEval2012_Task6}     & 10.9 \\
STS13 \citep{SemEval2013_SharedTask}& 8.9 \\
STS14 \citep{SemEval2014_Task10}    & 9.2 \\
STS15 \citep{SemEval2015_Task2}     & 10.6 \\
STS17 \citep{SemEval2017_Task1}     & 8.7 \\
STS22 \citep{SemEval2022_Task8}     & 461.0 \\
STSB \citep{SemEval2017_Task1}      & 9.9 \\
\midrule
\textbf{Average} & 61.1 \\
\bottomrule
\end{tabularx}
\caption{Average word count per input text for English STS datasets.}
\label{tab:00_wordcount_sts_results}
\end{center}
\end{table}
%
%
% ##########################################################################################
% ############################ METHOD ######################################################
% ##########################################################################################
\section{Method}%
% ------------------------------------------
% ---------- FIGURE METHOD-- ---------------
% ------------------------------------------
\begin{figure*}[!ht]
\begin{center}
  \includegraphics[width=\linewidth]{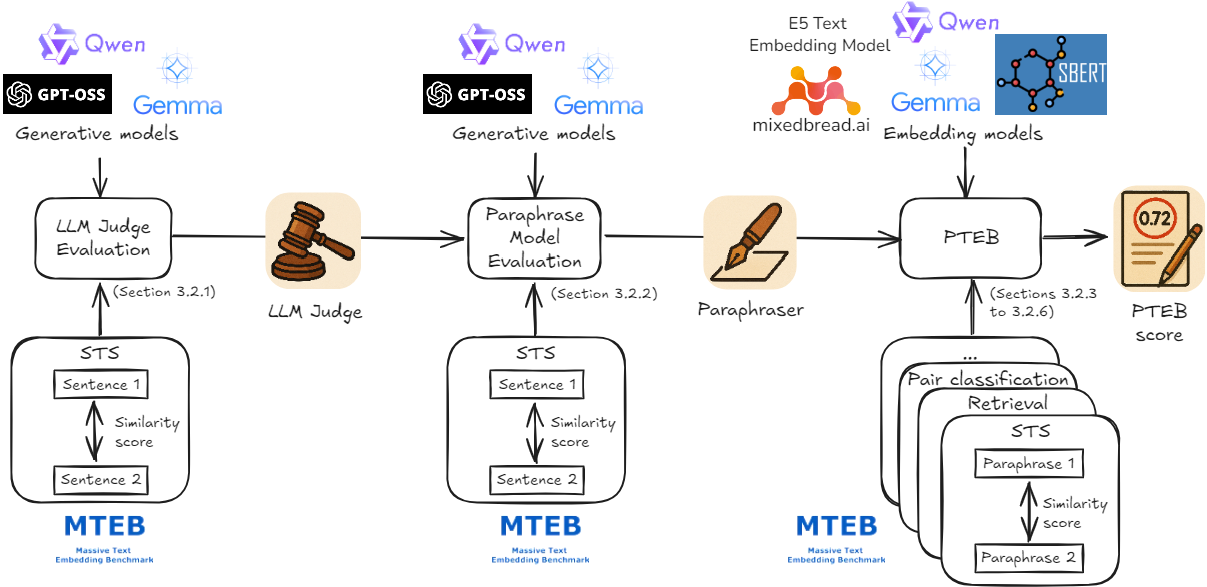}
  \caption{MTEB STS data is used to evaluate different generative LLMs for their STS rating capability. Using the selected LLM judge, generative LLMs are evaluated for paraphrasing MTEB STS data. Finally, the best paraphrase model is used to paraphrase MTEB datasets at eval-time generating a dynamic benchmark for embedding models.}
  \label{fig:method}
\end{center}
\end{figure*}
% ------------------------------------
Our method is illustrated in \autoref{fig:method}. We apply generative LLMs in two distinct roles: First, as an LLM judge to select a paraphrase model. Second, as a paraphraser generating paraphrases during evaluation. To evaluate and select the LLM judge, we use sentence pairs from the MTEB STS datasets. Without requiring human evaluations or direct validation of the generated paraphrases, our method anchors PTEB in scientific rigor by grounding it in STS gold scores. \autoref{tab:00_wordcount_sts_results} lists the STS datasets that we use and shows that the STS datasets cover a range of different text lengths ranging from short sentences to STS22's paragraphs with an average length of 461 words. Throughout this paper, cosine similarity is applied as the similarity metric between embeddings. Generally, we use the same metrics as MTEB, for example, Spearman's rank correlation for STS (see \autoref{tab:pteb_datasets} in \autoref{sec:app_datasets} for a list of all metrics per task/dataset). All averages are reported as macro-averages; for robust effect size estimation we additionally calculate the Hodges-Lehmann (HL) estimator~\citep{HodgesLehmann_HL-Estimator1963} with 95\% confidence intervals (CI) for selected analyses.\footnote{The HL estimator is a robust estimator calculated as the median of all Walsh averages, i.e., the pairwise means formed from every self-pair and unique unordered pair. To estimate its CIs we use a percentile bootstrap with 1000 iterations.}

\textbf{1. LLM Judge Evaluation.}
The MTEB STS datasets provide similarity ratings on bounded scales, obtained through annotation methods like Amazon Mechanical Turk. We evaluate generative LLMs based on their ability to assess the semantic similarity of sentence pairs vs. these STS gold ratings in order to select an LLM judge model.
%
%  -----------------------------------------------------
%  ---------- PERFORMANCE VS GOLD LABELS ---------------
%  -----------------------------------------------------
\begin{figure*}[!ht]
\begin{center}
  \includegraphics[width=\linewidth]{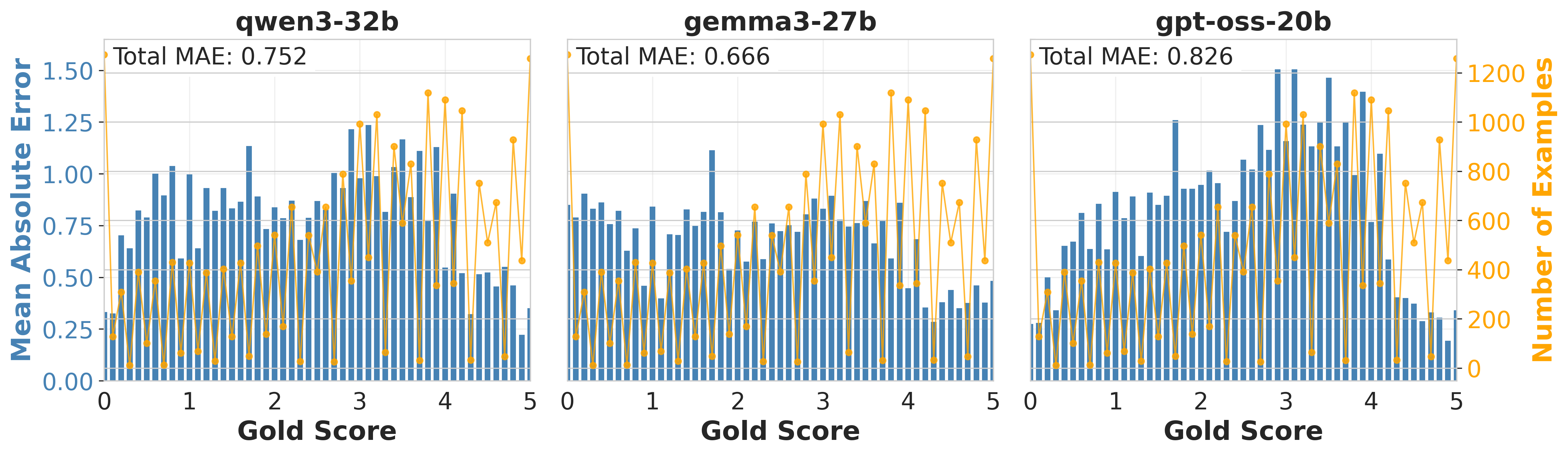}
  \caption{LLM judge mean average error (MAE) and number of examples per gold score on MTEB STS.}
  \label{fig:performance_vs_gold_labels_by_model}
\end{center}
\end{figure*}

\textbf{2. Paraphrase Model Evaluation.}
Next, we apply generative LLMs to paraphrase the MTEB STS datasets. The LLM judge selected in the previous step rates the semantic similarity of the paraphrases on a scale from 0 to 5. Based on semantic similarity, (normalised) edit distance (ED)\footnote{All reported EDs have been normalised by the maximal sentence length: $ED(s_1,s_2) = ED_{raw}(s_1,s_2)/max(len(s_1),len(s_2)))$ with $s_1$ and $s_2$ being two sentences, $len(s)$ the word count of sentence $s$, and $ED_{raw}$ the non-normalised ED.}, and runtime, we select the paraphrase model ("paraphraser"). We include the ED as a criterion to avoid textually overly similar paraphrases compared to the original text (larger ED is better).

\textbf{3. Embedding Model Evaluation on PTEB.}
Finally, we apply the selected paraphraser to rephrase MTEB datasets at eval-time generating a dynamic benchmark. This allows us to evaluate various embedding models on the paraphrases and compare their scores with the original MTEB datasets. For details on all datasets incl. the metrics used see \autoref{tab:pteb_datasets} in \autoref{sec:app_datasets}. We run PTEB using $n=6$ different paraphrasing seeds since single‐run estimates can be unreliable~\citep{Reimers2017_ReportingScoreDistributions,Reimers2018,Dror2019}. 

Given that steps 1 and 2 of our method rely on English datasets, we validate a subset of the paraphrases generated by the selected paraphrase model for non-English datasets using human raters. For selected languages, we randomly select 50 sentence pairs each and collect 2 human ratings per pair. We established an a priori acceptability threshold of $\geq 3.00$ for the mean human similarity ratings (scale: 0 to 5). See \autoref{sec:app_human_eval} for further details.
%
% ##########################################################################################
% ############################ EXPERIMENTS #################################################
% ##########################################################################################
\section{Experiments}%
% ################################################
% ########## EXPERIMENTS #########################
% ################################################
In this section, we present the results of our experiments. We begin by outlining the implementation details, followed by the results of the LLM judge evaluation. We then examine how the selected LLM judge rates various generative models for paraphrasing and use the best paraphraser for PTEB to evaluate various embedding models across STS and non-STS tasks as well as non-English languages. Finally, we validate PTEB using human evaluation and a prompt sensitivity analysis.  

\subsection{Implementation}
For the LLM judge and paraphraser selection, we assessed 3 generative open-weights models: \texttt{gemma~3-27b}~\citep{gemma3}, \texttt{gpt-oss-20b}~\citep{gpt-oss}, and \texttt{qwen3-32b}~\citep{qwen3}. Our selection criteria were: (1) open weights to enable reproducibility and auditability; (2) strong general reasoning capabilities based on public benchmarks; (3) moderate size (20-32B parameters) to run on consumer hardware with quantisation; and (4) availability via widely-used inference frameworks (Ollama, vLLM). These constraints ensure that a wide range of researchers and practitioners can reproduce and extend PTEB. Implementation details including hyperparameters can be found in \autoref{sec:app_implementation_details}.
\begin{table}[!ht]
\begin{center}
\small
\begin{tabularx}{\columnwidth}{XXXX}
\toprule
\textbf{Dataset} & \textbf{\mbox{gemma3-27b}} & \textbf{\mbox{gpt-oss-20b}} & \textbf{\mbox{qwen3-32b}} \\
\midrule
BIOSSES & \textbf{88.53} & 83.75 & 80.94 \\
SICK-R  & \textbf{76.61} & 70.84 & 72.21 \\
STS12   & \textbf{75.13} & 66.95 & 70.42 \\
STS13   & \textbf{89.12} & 87.07 & 87.60 \\
STS14   & \textbf{83.82} & 81.95 & 81.55 \\
STS15   & \textbf{89.87} & 87.26 & 86.70 \\
STS17   & \textbf{91.42} & 88.99 & 87.53 \\
STS22   & 66.87 & 68.82 & \textbf{70.59} \\
STSB   & \textbf{87.26} & 84.50 & 85.67 \\
\midrule
\textbf{Average} & \textbf{83.18} & 80.02 & 80.89 \\
\midrule
\end{tabularx}
\caption{Spearman's rank correlation between LLM Judge similarity scores and gold ratings for STS datasets; best scores bold. (in \%)}
\label{tab:01_judge_sts_results}
\end{center}
\end{table}
As encoders, we evaluated \texttt{qwen3-8b}~\citep{qwen3}, \texttt{mxbai-embed-large-v1}~\citep{mxbai_large}, \texttt{e5-mistral-7b-instruct}~\citep{Wang2024a, Wang2024b}, and \texttt{embeddinggemma-300m}~\citep{EmbeddingGemma2025} as SOTA encoders. As a baseline, we included the Sentence Transformer \texttt{all-mpnet-base-v2}~\citep{Reimers2019}. These models were selected to include both models that currently lead or rank high on the MTEB leaderboard and a baseline model with lower performance. To ensure consistency between original and paraphrased conditions, the evaluation pipeline was re-implemented following the MTEB specification in \citet{Muennighoff2023}.\footnote{Our scripts are not identical with the current MTEB-evaluation but follow the original paper, e.g. MTEB now undersamples classification data to include 8 examples per label instead of using the complete train/test splits.} For hyperparameters see \autoref{sec:app_implementation_details}; for all prompts \autoref{sec:app_prompts}.
% ################################################
% ############ EXPERIMENTAL RESULTS ##############
% ################################################
\subsection{Experimental Results}
% 
%  ------------------------------------------------------------------
%  -------------- TABLE LLM JUDGE SCORES STS ........----------------
%  ------------------------------------------------------------------
%
% 
% --------------- LLM JUDGE EVAL -------------------
% 
\subsubsection{LLM Judge Evaluation}
\label{sec:llm_judge_eval}
\begin{figure}[!ht]
\begin{center}
  \includegraphics[width=\linewidth]{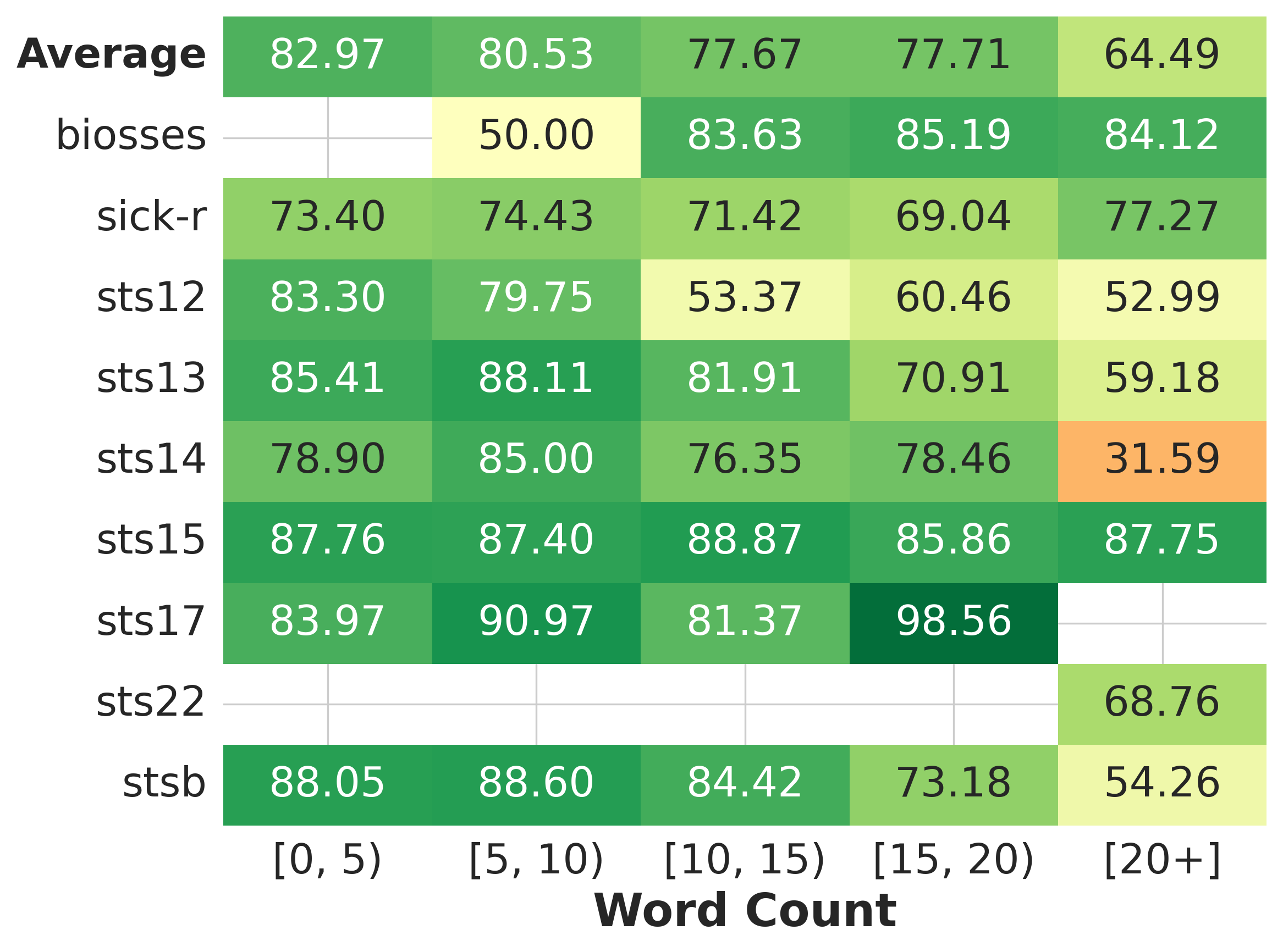}
  \caption{Spearman’s rank correlation between LLM judge similarity scores and gold ratings for different word count bins; empty cells indicate that the dataset does not contain sentences with that word count. (in \%)}
  \label{fig:judge_word_intervals}
\end{center}
\end{figure}

Analysing the mean absolute error (MAE) relative to the gold rating distribution (that is, the similarity scores) shows that all models have lower errors on sentence pairs with high ($\geq 4$) similarity scores (\autoref{fig:performance_vs_gold_labels_by_model}). However, while Qwen3 and GPT-OSS also have lower MAEs on sentence pairs with labels close to 0, this does not hold for the Gemma-3 model.
 \autoref{tab:01_judge_sts_results} presents the results for each LLM judge per dataset. Overall, \texttt{gemma3-27b} achieved the highest performance on MTEB STS. With the exception of STS22 the model outperformed GPT-OSS and Qwen3 on all datasets.
Examining performance by dataset and various word count bins (averaged over all 3 LLM judges), \autoref{fig:judge_word_intervals} shows that sentences with $\geq 20$ words yield the lowest performance (average score 64.49\%), while sentences with $<10$ words achieve an average Spearman's rank correlation of $\geq 80\%$. In particular, on long sentence pairs in STS14, STS12, and STSB the LLM judges score only 31.59\%, 52.99\% and 54.26\%.
% --------------------------------------------
%  ---------- WORD BIN HEATMAP ---------------
% --------------------------------------------
% 
% 
%  ------------------------------------------------------------------
%  -------------- TABLE LLM JUDGE SCORES WORD COUNTS ----------------
%  ------------------------------------------------------------------

To assess the effect of model size, we performed an ablation by varying the number of parameters (\autoref{tab:01_llm_judge_ablation}). Gemma-3 12B performs comparably to the 27B model (82.76\% vs. 83.18\%). By contrast, the 4B and, especially, the 1B and 270M variants exhibit substantial performance drops (72.74\%, 33.28\%, and 1.9\% respectively).
\begin{table}[!ht]
\small
\begin{center}
\begin{tabularx}{\columnwidth}{XXXXX}
\toprule
\multicolumn{5}{c}{\textbf{STS Judge Model Size}} \\ 
\midrule
\textbf{270M} & \textbf{1B} & \textbf{4B} & \textbf{12B} & \textbf{27B} \\
\midrule
1.9\% & 33.28\% & 72.74\% & 82.76\% & 83.18\% \\
\bottomrule
\end{tabularx}
\caption{LLM judge ablation (parameter count) for Gemma-3 grading semantic similarity on STS datasets.}
\label{tab:01_llm_judge_ablation}
\end{center}
\end{table}
\subsubsection{Paraphrase Model Evaluation}
\label{sec:paraphrase_model_eval}
\begin{figure}[!ht]
\begin{center}
  \includegraphics[width=\linewidth]{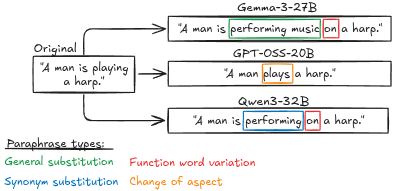}
  \caption{Paraphrase examples (types based on \citet{Bhagat2013_Paraphrase_Types}).}
  \label{fig:paraphrase_examples}
\end{center}
\end{figure}

% -----------------------------------------------------------------------------
% ---------------- Praphrase Model Selection ----------------------------------
% -----------------------------------------------------------------------------
Using \texttt{gemma3-27b} as the LLM judge, we observe that also as a paraphraser it not only achieves semantic similarity scores close to the model with the highest similarity score, \texttt{gpt-oss-20b}, but produces paraphrases with larger ED at a lower runtime (\autoref{tab:02_paraphrase_evaluation}). The larger ED is illustrated by the examples shown in \autoref{fig:paraphrase_examples}. Hence, overall, \texttt{gemma3-27b} is the most suitable paraphraser for PTEB.

To mitigate potential within-model bias arising from using \texttt{gemma3-27b} as both the paraphraser and the LLM judge, we validated the semantic similarity ratings using \texttt{gpt-oss-20b} and \texttt{qwen3-32b} as LLM judges (\autoref{tab:02_paraphrase_cross_eval}). Additionally, we added \texttt{mistral-small3.2:24b} as a held-out judge. All non-Gemma LLM judges confirm high similarity scores for Gemma's paraphrases. 
% ----------------------------------------------------------------------------
% ------------ TABLE: AVG. PARAPHRASE SCORES AND EDIT DISTANCES -------------- 
% ----------------------------------------------------------------------------
% 
% 
\begin{table}[!ht]
\small
\centering
\begin{tabularx}{\columnwidth}{lccc}
\toprule
\makecell[l]{\textbf{Paraphrase}\\\textbf{Model}} &
\makecell{\textbf{Semantic}\\\textbf{Similarity}} &
\makecell{\textbf{Edit}\\\textbf{Distance}} &
\makecell{\textbf{Run-}\\\textbf{time}} \\
\midrule
gemma3-27b & 4.26 & \textbf{0.81} & \textbf{0.27} \\
gpt-oss-20b & \textbf{4.44} & 0.55 & 1.24 \\
qwen3-32b & 4.39 & 0.63 & 5.35 \\
\bottomrule
\end{tabularx}
\caption{Paraphrase evaluation results showing average semantic similarity (0--5 scale) and normalized edit distance compared to originals (in \%), as well as runtime (seconds per text) on STS datasets; best results bold.}
\label{tab:02_paraphrase_evaluation}
\end{table}
%  ---------- FIGURE EXAMPLES ---------------
\begin{table}[!ht]
\small
\centering
\begin{tabularx}{\columnwidth}{
  >{\raggedright\arraybackslash}X 
  c
  c
}
\toprule
\textbf{LLM Judge} & \textbf{Average} & \textbf{HL Estimator [CI]} \\
\midrule
gemma3-27b & 4.26 & 4.24 [4.22, 4.30] \\
gpt-oss-20b & 4.68 & 4.68 [4.65, 4.72] \\
qwen3-32b & 4.52 & 4.52 [4.50, 4.54] \\
mistral-small3.2-24b & 4.22 & 4.22 [4.17, 4.27] \\
\bottomrule
\end{tabularx}
\caption{Semantic similarity ratings (average and Hodges-Lehmann (HL) Estimator with 95\% confidence intervals) between original texts and paraphrases generated with \texttt{gemma--27b} using different LLMs as judges.}
\label{tab:02_paraphrase_cross_eval}
\end{table}
\subsubsection{PTEB for STS}
\begin{table*}[!t]
\small
\centering
\begin{tabularx}{\textwidth}{
  >{\arraybackslash}X
  >{\centering\arraybackslash}p{0.0505\textwidth}
  >{\centering\arraybackslash}p{0.0505\textwidth}
  >{\centering\arraybackslash}p{0.0505\textwidth}
  >{\centering\arraybackslash}p{0.0505\textwidth}
  >{\centering\arraybackslash}p{0.0505\textwidth}
  >{\centering\arraybackslash}p{0.0505\textwidth}
  >{\centering\arraybackslash}p{0.0505\textwidth}
  >{\centering\arraybackslash}p{0.0505\textwidth}
  >{\centering\arraybackslash}p{0.0505\textwidth}
  >{\centering\arraybackslash}p{0.0505\textwidth}
  >{\centering\arraybackslash}p{0.075\textwidth}}
\toprule
\textbf{Dataset} 
  & \multicolumn{2}{c}{\makecell{\textbf{all-mpnet-}\\\textbf{base-v2}}}
  & \multicolumn{2}{c}{\makecell{\textbf{embedding}\\\textbf{gemma-300m}}}
  & \multicolumn{2}{c}{\makecell{\textbf{mxbai-embed-}\\\textbf{large-v1}}}
  & \multicolumn{2}{c}{\makecell{\textbf{e5-mistral-}\\\textbf{7b-instruct}}}
  & \multicolumn{2}{c}{\makecell{\textbf{qwen3-em-}\\\textbf{bedding-8b}}}
  & \textbf{Average} \\
\cmidrule(lr{0.6em}){2-3}\cmidrule(lr{0.6em}){4-5}\cmidrule(lr{0.6em}){6-7}\cmidrule(lr{0.6em}){8-9}\cmidrule(lr{0.6em}){10-11}
 & \textbf{Orig.} & \textbf{PTEB} 
 & \textbf{Orig.} & \textbf{PTEB} 
 & \textbf{Orig.} & \textbf{PTEB} 
 & \textbf{Orig.} & \textbf{PTEB} 
 & \textbf{Orig.} & \textbf{PTEB} \\ 
\midrule
BIOSSES 
& 80.39 & 77.22
& 70.27 & 64.67
& 86.09 & \textbf{84.06}
& 84.61 & 77.68
& \textbf{86.29} & 81.44 \\
& \multicolumn{2}{c}{$\Delta$\red{-3.17} / ±2.23} 
& \multicolumn{2}{c}{$\Delta$\red{-5.60} / ±1.61} 
& \multicolumn{2}{c}{$\Delta$\red{-2.03} / ±1.11} 
& \multicolumn{2}{c}{$\Delta$\red{-6.93} / ±1.01} 
& \multicolumn{2}{c}{$\Delta$\red{-4.85} / ±1.53} 
& $\Delta$\red{-4.52}\\
\midrule
SICK-R  
& 80.60 & 77.45
& 71.99 & 70.28
& 82.79 & 78.24
& 80.76 & 79.01
& \textbf{84.97} & \textbf{81.88} \\
& \multicolumn{2}{c}{$\Delta$\red{-3.15} / ±0.09} 
& \multicolumn{2}{c}{$\Delta$\red{-1.71} / ±0.07} 
& \multicolumn{2}{c}{$\Delta$\red{-4.55} / ±0.07} 
& \multicolumn{2}{c}{$\Delta$\red{-1.75} / ±0.05} 
& \multicolumn{2}{c}{$\Delta$\red{-3.09} / ±0.07}
& $\Delta$\red{-2.85} \\
\midrule
STS12   
& 72.64 & 61.85
& 63.53 & 54.95
& 79.07 & 67.91
& 75.83 & \textbf{68.23}
& \textbf{81.41} & 67.21 \\
& \multicolumn{2}{c}{$\Delta$\red{-10.79} / ±0.29} 
& \multicolumn{2}{c}{$\Delta$\red{-8.58} / ±0.41} 
& \multicolumn{2}{c}{$\Delta$\red{-11.16} / ±0.24} 
& \multicolumn{2}{c}{$\Delta$\red{-7.60} / ±0.32} 
& \multicolumn{2}{c}{$\Delta$\red{-14.20} / ±0.30}
& $\Delta$\red{-10.47} \\
\midrule
STS13   
& 83.49 & 78.91
& 67.71 & 73.70
& \textbf{89.81} & 84.19
& 84.31 & \textbf{84.35}
& 87.89 & 82.90 \\
& \multicolumn{2}{c}{$\Delta$\red{-4.58} / ±0.56} 
& \multicolumn{2}{c}{$\Delta$\green{+5.99} / ±0.64} 
& \multicolumn{2}{c}{$\Delta$\red{-5.62} / ±0.33} 
& \multicolumn{2}{c}{$\Delta$\green{+0.04} / ±0.35} 
& \multicolumn{2}{c}{$\Delta$\red{-4.99} / ±0.41}
& $\Delta$\red{-1.83} \\
\midrule
STS14   
& 78.00 & 72.14
& 64.95 & 65.03
& \textbf{85.22} & \textbf{78.41}
& 79.26 & 78.12
& 83.32 & 77.10 \\
& \multicolumn{2}{c}{$\Delta$\red{-5.86} / ±0.26} 
& \multicolumn{2}{c}{$\Delta$\green{+0.08} / ±0.19} 
& \multicolumn{2}{c}{$\Delta$\red{-6.81} / ±0.12} 
& \multicolumn{2}{c}{$\Delta$\red{-1.14} / ±0.21} 
& \multicolumn{2}{c}{$\Delta$\red{-6.22} / ±0.22}
& $\Delta$\red{-3.99} \\
\midrule
STS15   
& 85.66 & 81.58
& 77.19 & 74.47
& 89.34 & 85.05
& 85.20 & 86.20
& \textbf{89.56} & \textbf{86.50} \\
& \multicolumn{2}{c}{$\Delta$\red{-4.08} / ±0.11} 
& \multicolumn{2}{c}{$\Delta$\red{-2.72} / ±0.19} 
& \multicolumn{2}{c}{$\Delta$\red{-4.29} / ±0.20} 
& \multicolumn{2}{c}{$\Delta$\green{+1.00} / ±0.10} 
& \multicolumn{2}{c}{$\Delta$\red{-3.06} / ±0.12}
& $\Delta$\red{-2.63} \\
\midrule
STS17   
& 90.61 & 85.06
& 82.61 & 75.28
& 89.22 & 85.00
& 88.86 & 85.61
& \textbf{92.19} & \textbf{88.60}\\
& \multicolumn{2}{c}{$\Delta$\red{-5.55} / ±0.25} 
& \multicolumn{2}{c}{$\Delta$\red{-7.33} / ±0.70} 
& \multicolumn{2}{c}{$\Delta$\red{-4.22} / ±0.28} 
& \multicolumn{2}{c}{$\Delta$\red{-3.25} / ±0.31} 
& \multicolumn{2}{c}{$\Delta$\red{-3.59} / ±0.21}
& $\Delta$\red{-4.79}  \\
\midrule
STS22   
& 68.31 & 69.39
& 58.05 & 63.21
& 69.02 & 68.97
& 69.66 & 67.49
& \textbf{69.98} & 68.61 \\
& \multicolumn{2}{c}{$\Delta$\green{+1.08} / ±0.48} 
& \multicolumn{2}{c}{$\Delta$\green{+5.16} / ±0.67} 
& \multicolumn{2}{c}{$\Delta$\red{-0.05} / ±0.38} 
& \multicolumn{2}{c}{$\Delta$\red{-2.17} / ±0.74} 
& \multicolumn{2}{c}{$\Delta$\red{-1.37} / ±0.66}
& $\Delta$\green{+0.53} \\
\midrule
STSB    
& 83.43 & 74.90
& 67.45 & 66.97
& \textbf{89.29} & 82.20
& 84.58 & 82.02
& 88.50 & \textbf{84.56} \\
& \multicolumn{2}{c}{$\Delta$\red{-8.53} / ±0.35} 
& \multicolumn{2}{c}{$\Delta$\red{-0.48} / ±0.53} 
& \multicolumn{2}{c}{$\Delta$\red{-7.09} / ±0.36} 
& \multicolumn{2}{c}{$\Delta$\red{-2.56} / ±0.38} 
& \multicolumn{2}{c}{$\Delta$\red{-3.94} / ±0.32}
& $\Delta$\red{-4.52} \\
\midrule
\textbf{Average} 
& 80.35 & 75.39
& 69.31 & 67.62
& 84.43 & 79.34
& 81.45 & 78.75
& \textbf{84.90} & \textbf{79.87} \\
& \multicolumn{2}{c}{$\Delta$\red{-4.96}}
& \multicolumn{2}{c}{$\Delta$\red{-1.69}}
& \multicolumn{2}{c}{$\Delta$\red{-5.09}}
& \multicolumn{2}{c}{$\Delta$\red{-2.70}}
& \multicolumn{2}{c}{$\Delta$\red{-5.03}}
& $\Delta$\red{-3.90} \\
\bottomrule
\end{tabularx}
\caption{Embedding model performance on original and PTEB STS datasets. $\Delta$ denotes the difference with \red{Red} indicating performance decreases and \green{Green} improvements. Sample standard deviation over the different paraphrases denoted by ±. Bold marks row-wise best for original and paraphrased text separately. ($n=6$, in \%)}
\label{tab:03_sts_embedding_results}
\end{table*}
\begin{table}[!ht]
\small
\centering
\begin{tabularx}{\columnwidth}{l c c}
\toprule
\textbf{all-mpnet-base-v2 vs.}  & \textbf{$\Delta_{HL}$ [CI]} & \textbf{$p_{Holm}$} \\
\midrule
\multirow{2}{*}{embedding-gemma-300m} & +7.33 & 6.50e-10 \\
                                      & [7.02, 8.00] & \\
\midrule
\multirow{2}{*}{mxbai-embed-large-v1} & -3.74 & 1.33e-08 \\
                                      & [-5.00, -3.14] & \\
\midrule
\multirow{2}{*}{e5-mistral-7b-instruct} & -3.48 & 1.40e-07 \\
                                        & [-4.37, -2.56] & \\
\midrule
\multirow{2}{*}{qwen3-embedding-8b} & -4.53 & 1.48e-09 \\
                                    & [-4.82, -4.19] & \\
\bottomrule
\end{tabularx}
\caption{Comparison of encoders on PTEB STS vs. \texttt{all-mpnet-base-v2} as baseline. $\Delta_{HL}$ is the Hodges-Lehmann estimator of the location shift with 95\% CI; negative values favour the listed model. P-values are Holm-adjusted. Based on $9\times6$ dataset-run pairs.}
\label{tab:03_stat_analysis}
\end{table}
On average, all models decline on PTEB vs. the original STS datasets with the smallest drop for \texttt{embeddinggemma-300m} (\autoref{tab:03_sts_embedding_results}). \texttt{e5-mistral-7b-instruct} decreases by 2.70\%, while others drop by up to 5.09\%. The gap is largest on STS12 (\mbox{-10.47}) while STS22 shows a small average improvement of +0.53.
\begin{table}[!ht]
\small
\begin{center}
\begin{tabularx}{\columnwidth}{lXXX}
\toprule
                          & \multicolumn{3}{c}{\textbf{Embedding Model Size}} \\ 
\cmidrule{2-4}
                          & \textbf{0.6B} & \textbf{4B} & \textbf{8B} \\
\midrule
    Original STS Score    & 81.42         & 84.26       & \textbf{84.90}\\
    PTEB STS Score        & 77.19         & 78.72       & \textbf{79.87}\\
\midrule
    Runtime [s]           & \textbf{135}  & 201         & 263 \\
    VRAM [GB]             & \textbf{1.1}  & 7.6         & 14.1 \\
\bottomrule
\end{tabularx}
\caption{Qwen3 embedding ablation on the numbers of parameters for STS; STS scores in \%. Runtime is the time for encoding only. VRAM refers to memory usage of the model only. Best in bold. ($n=6$ runs)}
\label{tab:03_ablation_emb_models}
\end{center}
\end{table}
\begin{table*}[!t]
\small
\centering
\begin{tabularx}{\textwidth}{
  >{\arraybackslash}X
  >{\centering\arraybackslash}p{0.050\textwidth}
  >{\centering\arraybackslash}p{0.050\textwidth}
  >{\centering\arraybackslash}p{0.050\textwidth}
  >{\centering\arraybackslash}p{0.050\textwidth}
  >{\centering\arraybackslash}p{0.050\textwidth}
  >{\centering\arraybackslash}p{0.050\textwidth}
  >{\centering\arraybackslash}p{0.050\textwidth}
  >{\centering\arraybackslash}p{0.050\textwidth}
  >{\centering\arraybackslash}p{0.050\textwidth}
  >{\centering\arraybackslash}p{0.050\textwidth}
  >{\centering\arraybackslash}p{0.059\textwidth}}
\toprule
\textbf{Task} 
  & \multicolumn{2}{c}{\makecell{\textbf{all-mpnet-}\\\textbf{base-v2}}}
  & \multicolumn{2}{c}{\makecell{\textbf{embedding}\\\textbf{gemma-300m}}}
  & \multicolumn{2}{c}{\makecell{\textbf{mxbai-embed-}\\\textbf{large-v1}}}
  & \multicolumn{2}{c}{\makecell{\textbf{e5-mistral-}\\\textbf{7b-instruct}}}
  & \multicolumn{2}{c}{\makecell{\textbf{qwen3-em-}\\\textbf{bedding-8b}}}
  & \textbf{Average} \\
\cmidrule(lr{0.6em}){2-3}\cmidrule(lr{0.6em}){4-5}\cmidrule(lr{0.6em}){6-7}\cmidrule(lr{0.6em}){8-9}\cmidrule(lr{0.6em}){10-11}
 & \textbf{Orig.} & \textbf{PTEB} 
 & \textbf{Orig.} & \textbf{PTEB} 
 & \textbf{Orig.} & \textbf{PTEB} 
 & \textbf{Orig.} & \textbf{PTEB} 
 & \textbf{Orig.} & \textbf{PTEB}
 & \\
\midrule
\textbf{STS} 9 datasets 
   & 80.35 & 75.39
   & 69.31 & 67.62
   & 84.43 & 79.34
   & 81.45 & 78.75
   & \textbf{84.90} & \textbf{79.87}
   & \\
(see \autoref{tab:03_sts_embedding_results})& \multicolumn{2}{c}{$\Delta$\red{-4.96} / ±1.37}
& \multicolumn{2}{c}{$\Delta$\red{-1.69} / ±0.49}
& \multicolumn{2}{c}{$\Delta$\red{-5.09} / ±0.40}
& \multicolumn{2}{c}{$\Delta$\red{-2.70} / ±0.43}
& \multicolumn{2}{c}{$\Delta$\red{-5.03} / ±0.42}
& $\Delta$\red{-3.89} \\
\midrule
\textbf{Classification}  
   & 92.23 & 85.90
   & 85.83 & 81.16
   & \textbf{93.66} & \textbf{87.86}
   & 90.99 & 84.90
   & 92.85 & 86.77
   & \\
Banking77 & \multicolumn{2}{c}{$\Delta$ \red{-6.33} / ±0.48}
& \multicolumn{2}{c}{$\Delta$ \red{-4.67} / ±0.37}
& \multicolumn{2}{c}{$\Delta$ \red{-5.80} / ±0.39}
& \multicolumn{2}{c}{$\Delta$ \red{-6.09} / ±0.43}
& \multicolumn{2}{c}{$\Delta$ \red{-6.08} / ±0.37}
& $\Delta$\red{-5.79} \\
\midrule
\textbf{Clustering}  
   & 50.18 & 47.44
   & 32.99 & 34.51
   & 51.25 & 48.52
   & 45.61 & 39.57
   & \textbf{55.51} & \textbf{50.03}
   & \\
TwentyNewsGr. & \multicolumn{2}{c}{$\Delta$ \red{-2.74} / ±0.71}
& \multicolumn{2}{c}{$\Delta$ \green{+1.52} / ±0.51}
& \multicolumn{2}{c}{$\Delta$ \red{-2.73} / ±0.56}
& \multicolumn{2}{c}{$\Delta$ \red{-6.04} / ±0.99}
& \multicolumn{2}{c}{$\Delta$ \red{-5.48} / ±0.75}
& $\Delta$\red{-3.09} \\
\midrule
\textbf{Pair Classif.}  
   & 73.85 & 71.65
   & 59.03 & 63.90
   & \textbf{78.54} & 73.65
   & 75.36 & 74.40
   & 72.22 & \textbf{74.21}
   & \\  
TwitterSemEval & \multicolumn{2}{c}{$\Delta$ \red{-2.20} / ±0.39}
& \multicolumn{2}{c}{$\Delta$ \green{+4.87} / ±0.10}
& \multicolumn{2}{c}{$\Delta$ \red{-4.89} / ±0.37}
& \multicolumn{2}{c}{$\Delta$ \red{-0.96} / ±0.49}
& \multicolumn{2}{c}{$\Delta$ \green{+1.99} / ±0.45}
& $\Delta$\red{-0.64} \\
\midrule
\textbf{Reranking}  
   & 65.81 & 64.30
   & 48.53 & 48.86
   & 65.20 & 63.78
   & 59.96 & 59.31
   & \textbf{67.55} & \textbf{64.98}
   & \\
AskUbuntuDupQ. & \multicolumn{2}{c}{$\Delta$ \red{-1.51} / ±0.40}
& \multicolumn{2}{c}{$\Delta$ \green{+0.33} / ±0.53}
& \multicolumn{2}{c}{$\Delta$ \red{-1.42} / ±0.30}
& \multicolumn{2}{c}{$\Delta$ \red{-0.65} / ±0.46}
& \multicolumn{2}{c}{$\Delta$ \red{-2.57} / ±0.26}
& $\Delta$\red{-1.16} \\
\midrule
\textbf{Retrieval}  
   & 45.82 & 45.38
   & 29.65 & 33.05
   & 65.11 & 61.36
   & 53.29 & 49.54
   & \textbf{75.26} & \textbf{73.30}
   & \\
ArguAna& \multicolumn{2}{c}{$\Delta$ \red{-0.44} / ±0.28}
& \multicolumn{2}{c}{$\Delta$ \green{+3.40} / ±0.28}
& \multicolumn{2}{c}{$\Delta$ \red{-3.75} / ±0.20}
& \multicolumn{2}{c}{$\Delta$ \red{-3.75} / ±0.38}
& \multicolumn{2}{c}{$\Delta$ \red{-1.96} / ±0.23}
& $\Delta$\red{-1.30} \\
\midrule
\textbf{Summarisation}  
   & 26.25 & 23.55
   & 20.35 & 15.57
   & 35.55 & 33.02
   & 34.24 & 32.27
   & \textbf{38.87} & \textbf{32.95}
   & \\
SummEval & \multicolumn{2}{c}{$\Delta$ \red{-2.70} / ±0.65}
& \multicolumn{2}{c}{$\Delta$ \red{-4.78} / ±0.58}
& \multicolumn{2}{c}{$\Delta$ \red{-2.53} / ±0.46}
& \multicolumn{2}{c}{$\Delta$ \red{-1.97} / ±1.15}
& \multicolumn{2}{c}{$\Delta$ \red{-5.92} / ±1.27}
& $\Delta$\red{-3.58} \\
\midrule
\textbf{Average} 
   & 62.07 & 59.09
   & 49.38 & 49.24
   & 67.68 & 63.93
   & 62.99 & 59.82
   & \textbf{69.59} & \textbf{66.02}
   & \\
& \multicolumn{2}{c}{$\Delta$ \red{-2.98}}
& \multicolumn{2}{c}{$\Delta$ \red{-0.15}}
& \multicolumn{2}{c}{$\Delta$ \red{-3.74}}
& \multicolumn{2}{c}{$\Delta$ \red{-3.17}}
& \multicolumn{2}{c}{$\Delta$ \red{-3.58}}
& $\Delta$\red{-2.72} \\
\bottomrule
\end{tabularx}
\caption{Embedding model performance across STS and Non-STS tasks (English); the STS total (from Table \ref{tab:03_sts_embedding_results}) is included as the first row. $\Delta$ denotes differences, \red{Red} marks decreases and \green{green} improvements. ± denotes the standard deviation on PTEB. Bold marks row-wise best for Original and PTEB separately. ($n=6$ runs; in \%)}
\label{tab:03_combined_task_results}
\end{table*}
Also the standard deviation across runs differs substantially between datasets. On SICK-R and STS15 the sample standard deviation ($\sigma$) is at the lower end ($\sigma \leq 0.09$ and $\sigma \leq 0.20$ respectively) in contrast to BIOSSES ($\sigma \in [1.01, 2.23]$).   

To test statistical significance, we use the Wilcoxon signed-rank test, which is widely applicable in NLP~\citep{Dror2018}.\footnote{Following concerns that bootstrapping requires larger sample sizes \citep{Sogaard2014_WhatsPvalueNLP, Hesterberg2014_WhatTeachersShouldKnowAboutTheBootstrap}, we avoid it for significance testing.} Pooling over the $9\times6$ STS dataset-run pairs, the negative HL location shifts ($\Delta_{HL}$) and Holm-corrected~\citep{Holm1979}\footnote{The Holm-method adjusts p-values when testing multiple hypotheses. It sorts the $m$ hypotheses by their p-values ($p_i$) and then checks, starting with the lowest p-value, for a given significance level $\alpha$ if $\forall i \in \{1,\dots,m\}: p_i < \frac{\alpha}{m-i+1}$~\citep{Calonico2025}).} p-values $\ll \alpha=0.05$ show that all embedding models except \texttt{embeddinggemma-300m} outperform the baseline significantly (see \autoref{tab:03_stat_analysis}; for a more conservative analysis averaging across runs before testing, see \autoref{tab:app_stat_analysis_aggregated} in  \autoref{sec:app_stat_run_agg_analysis}).
\subsubsection{Ablation on STS Embedding Model Size}
To analyse the influence of the size of the encoder, we conducted an ablation study using the Qwen3 models (\autoref{tab:03_ablation_emb_models}). The performance gap between the original and paraphrased datasets remains similar across all 3 Qwen3 variants, with the smallest model even displaying the narrowest gap.
% 
% 
% ------------------------- Non-STS Tasks -------------------  
% 
\subsubsection{PTEB for Non-STS Tasks}
We extend PTEB to 6 additional MTEB tasks in order to incorporate a broad range of tasks ranging from straightforward binary decisions to nuanced clustering of sentences. \autoref{tab:03_combined_task_results} shows that the average performance of all models drops on PTEB compared to original texts, although particularly \texttt{embeddinggemma-300m} improves on multiple tasks. \texttt{qwen3-8b} is the best-performing model in 5 of the 7 tasks on original data and 6 of 7 on PTEB. (For an additional visualisation see \autoref{sec:app_add_analyses}.)
% 
% 
% ------------- MULTILINGUAL PTEB
% 
% ----------------------------------------------------------------
% -------- MULTILINGUAL TABLE ------------------------------------
% ----------------------------------------------------------------
% 
% 
\begin{table*}[!t]
\small
\centering
\begin{tabularx}{\textwidth}{
  >{\arraybackslash}X
  >{\centering\arraybackslash}p{0.085\textwidth}
  >{\centering\arraybackslash}p{0.065\textwidth}
  >{\centering\arraybackslash}p{0.065\textwidth}
  >{\centering\arraybackslash}p{0.06\textwidth}
  >{\centering\arraybackslash}p{0.06\textwidth}
  >{\centering\arraybackslash}p{0.06\textwidth}
  >{\centering\arraybackslash}p{0.06\textwidth}
  >{\centering\arraybackslash}p{0.06\textwidth}
  >{\centering\arraybackslash}p{0.06\textwidth}}
\toprule
\textbf{Task} 
  & \multicolumn{2}{c}{\makecell{\textbf{paraphrase-multi-}\\\textbf{lingual-mpnet-base-v2}}}
  & \multicolumn{2}{c}{\makecell{\textbf{embedding}\\\textbf{gemma-300m}}}
  & \multicolumn{2}{c}{\makecell{\textbf{multilingual-e5-}\\\textbf{large-instruct}}}
  & \multicolumn{2}{c}{\makecell{\textbf{qwen3-em-}\\\textbf{bedding-8b}}}
  & \textbf{Average} \\
\cmidrule(lr{0.6em}){2-3}\cmidrule(lr{0.6em}){4-5}\cmidrule(lr{0.6em}){6-7}\cmidrule(lr{0.6em}){8-9}
 & \textbf{Orig.} & \textbf{PTEB} 
 & \textbf{Orig.} & \textbf{PTEB} 
 & \textbf{Orig.} & \textbf{PTEB} 
 & \textbf{Orig.} & \textbf{PTEB} 
 &  \\
\midrule
\textbf{Classification}  
   & 88.54 & 85.55
   & \textbf{90.51} & 86.11
   & 89.08 & 86.32
   & 89.83 & \textbf{87.04}
   &  \\
AmazonCounterf.
& \multicolumn{2}{c}{$\Delta$ \red{-2.99} / ±0.42}
& \multicolumn{2}{c}{$\Delta$ \red{-4.40} / ±0.36}
& \multicolumn{2}{c}{$\Delta$ \red{-2.77} / ±0.17}
& \multicolumn{2}{c}{$\Delta$ \red{-2.79} / ±0.31}
&  $\Delta$\red{-3.24}\\
\midrule
\textbf{Clustering}  
   & 18.98 & 19.41
   & 16.17 & 15.51
   & 24.77 & 21.04
   & \textbf{25.49} & \textbf{24.42}
   &  \\
MasakhaNEWS 
& \multicolumn{2}{c}{$\Delta$ \green{+0.43} / ±0.87}
& \multicolumn{2}{c}{$\Delta$ \red{-0.67} / ±1.43}
& \multicolumn{2}{c}{$\Delta$ \red{-3.72} / ±2.31}
& \multicolumn{2}{c}{$\Delta$ \red{-1.08} / ±1.53}
& $\Delta$\red{-1.26} \\
\midrule
\textbf{Pair Classifi-}  
   & \textbf{89.14} & \textbf{89.29}
   & 86.95 & 86.49
   & 87.62 & 87.23
   & 87.35 & 87.24
   &  \\  
\textbf{cation} RTE3 
& \multicolumn{2}{c}{$\Delta$ \green{+0.15} / ±0.05}
& \multicolumn{2}{c}{$\Delta$ \red{-0.46} / ±0.24}
& \multicolumn{2}{c}{$\Delta$ \red{-0.39} / ±0.13}
& \multicolumn{2}{c}{$\Delta$ \red{-0.11} / ±0.18}
& $\Delta$\red{-0.20}\\
\midrule
\textbf{Reranking}  
   & 58.74 & 58.93
   & 22.62 & 24.11
   & 69.51 & 69.80
   & \textbf{75.07} & \textbf{75.08}
   &  \\
RuBQReranking 
& \multicolumn{2}{c}{$\Delta$ \green{+0.19} / ±0.68}
& \multicolumn{2}{c}{$\Delta$ \green{+1.49} / ±0.60}
& \multicolumn{2}{c}{$\Delta$ \green{+0.29} / ±0.13}
& \multicolumn{2}{c}{$\Delta$ \green{+0.01} / ±0.30}
& $\Delta$\green{+0.50} \\
\midrule
\textbf{Retrieval}  
   & 58.26 & 57.78
   & 29.15 & 25.70
   & 72.62 & 73.16
   & \textbf{78.72} & \textbf{80.54}
   &  \\
TwitterHjerne  
& \multicolumn{2}{c}{$\Delta$ \red{-0.48} / ±1.21}
& \multicolumn{2}{c}{$\Delta$ \red{-3.45} / ±0.63}
& \multicolumn{2}{c}{$\Delta$ \green{+0.54} / ±1.18}
& \multicolumn{2}{c}{$\Delta$ \green{+1.82} / ±0.89}
&  $\Delta$\red{-0.40}\\
\midrule
\textbf{STS}  
   & \textbf{83.46} & 79.98
   & 61.89 & 59.74
   & 72.82 & 70.86
   & 83.35 & \textbf{80.24}
   &  \\
STS17  
& \multicolumn{2}{c}{$\Delta$ \red{-3.48} / ±0.19}
& \multicolumn{2}{c}{$\Delta$ \red{-2.15} / ±0.51}
& \multicolumn{2}{c}{$\Delta$ \red{-1.96} / ±0.26}
& \multicolumn{2}{c}{$\Delta$ \red{-3.11} / ±0.20}
&  $\Delta$\red{-2.67}\\
\midrule
\textbf{Average}  
   & 66.19 & 65.16
   & 51.22 & 49.61
   & 69.40 & 68.07
   & \textbf{73.30} & \textbf{72.43}
   &  \\
 & \multicolumn{2}{c}{$\Delta$ \red{-1.03}}
 & \multicolumn{2}{c}{$\Delta$ \red{-1.61}}
 & \multicolumn{2}{c}{$\Delta$ \red{-1.33}}
 & \multicolumn{2}{c}{$\Delta$ \red{-0.88}}
 &  $\Delta$\red{-1.21}\\
\bottomrule
\end{tabularx}
\caption{
Multilingial embedding model performance on original and PTEB datasets (25 languages); $\Delta$ denotes differences, \red{Red} marks decreases and \green{green} improvements. ± denotes the standard deviation on PTEB. Bold marks row-wise best for Original and PTEB separately. ($n=6$ runs; in \%)}
\label{tab:03_multilingual_results}
\end{table*}
\subsubsection{Multilingual PTEB}
We validate our results on 6 datasets from 6 different tasks that include non-English data (incl. the multilingual version of STS17) covering 25 languages. We examine both widely-studied languages (e.g. Arabic) and under-resourced ones (e.g. Danish) incl. 13 from sub-Saharan Africa (e.g. Suwaheli, Hausa, Amharic) to evaluate performance across linguistic boundaries (see \autoref{tab:03_multilingual_results} for results; see \autoref{sec:app_datasets} for the languages included).\footnote{We did not include a non-English dataset for the summarisation task, as MTEB does not provide one.} We used the same encoders as before if they are natively multilingual, or their multilingual variants alternatively. Consistent with English datasets, we find the largest drops on the classification task followed by STS, and clustering. An exception is the Russian RubQ dataset on which all models improve. As expected, for each task the average decreases are more pronounced on the English datasets compared to non-English datasets. 
% 
% ------------------------- Human Eval -------------------  
% 
\subsubsection{Human Evaluation}
\begin{figure}
    \centering
    \includegraphics[width=1.0\linewidth]{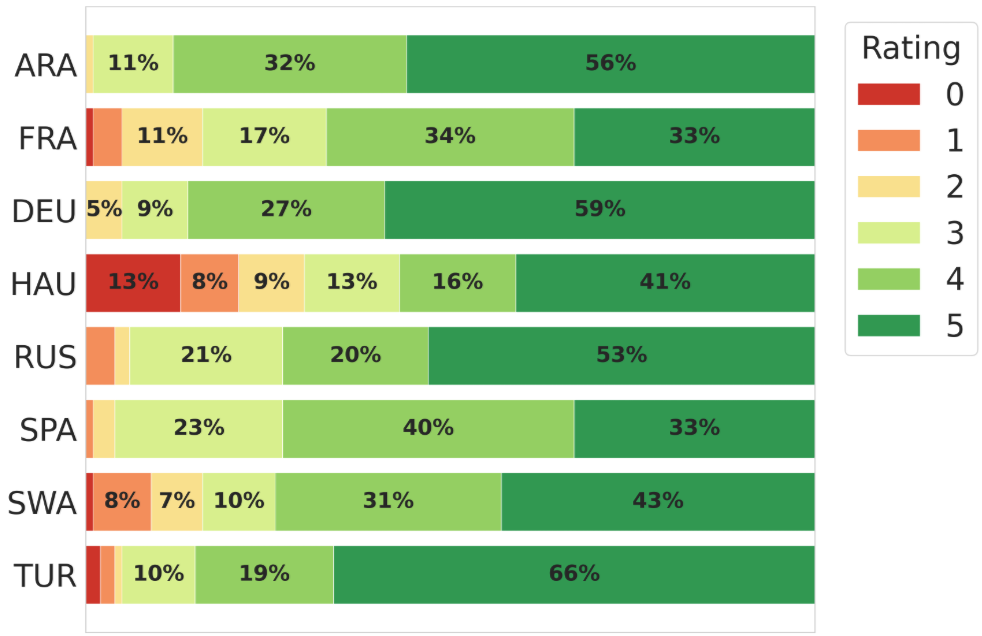}
    \caption{Human paraphrase similarity rating distribution (based on $n=2 \times 50$ ratings per language).}
    \label{fig:human_similarity_rating_distribution}
\end{figure}

Languages for human evaluation were selected to (1) cover high and low resource languages (including low-resource Sub-Saharan languages like Hausa and Swahili), (2) represent a wide range of PTEB tasks (spanning Classification, Clustering, Pair Classification, Reranking, and STS), and (3) encompass diverse language families (e.g., Afroasiatic, Indo-European, Niger-Congo, and Turkic). 
% 
% ----------------------------------------------------------------
% -------- HUMAN EVAL TABLE ------------------------------------
% ----------------------------------------------------------------
% 
\begin{table}[!ht]
\small
\centering
\begin{tabularx}{\columnwidth}{Xccc}
\toprule
\makecell[l]{\textbf{Language}} &
\makecell{\textbf{No. of}\\\textbf{Ratings}} &
\makecell{\textbf{Mean}\\\textbf{Similarity}} &
\makecell{\textbf{Median}\\\textbf{Similarity}} \\
\midrule
Arabic (ARA)  & 2x50 & 4.43 & 5.00 \\
French (FRA)  & 2x50 & 3.78 & 4.00 \\
German (DEU)  & 2x50 & 4.40 & 5.00 \\
Hausa (HAU)   & 2x50 & 3.34 & 4.00 \\
Russian (RUS) & 2x50 & 4.16 & 5.00 \\
Spanish (SPA) & 2x50 & 4.01 & 4.00 \\
Swahili (SWA) & 2x50 & 3.91 & 4.00 \\
Turkish (TUR) & 2x50 & 4.40 & 5.00 \\
\midrule
\textbf{Total} & \textbf{2x400} & \textbf{4.05} & \textbf{4.00} \\
\bottomrule
\end{tabularx}
\caption{Human evaluation of the similarity between original texts and paraphrases (on scale from 0 to 5).}
\label{tab:human_eval}
\end{table}
The human similarity ratings ranged from 3.34 (Hausa) to 4.43 (Arabic), with an average of 4.05 and a pooled median of 4.00 (\autoref{tab:human_eval}). This confirms a high degree of semantic similarity between the generated paraphrases and the original text. All languages exceeded the mean acceptability threshold. While the majority of sentence pairs were rated 3.00 or higher, 13\% of sentence pairs in Hausa received a similarity score of 0.00 (\autoref{fig:human_similarity_rating_distribution}).
% 
% --------------------------------------------------------------
% -------- HUMAN EVAL IMAGE ------------------------------------
% --------------------------------------------------------------
% 
% ------------------------- Prompt Sensitivity -------------------  
% 
\subsubsection{STS Paraphrase Prompting Sensitivity}
\label{sec:sts_prompt_sensitivity}
Since LLMs can be sensitive to prompt changes~\citep{Mizrahi2024}, we tested the influence of 3 variations of the paraphrase prompt and compared the PTEB STS scores and the ED between original and paraphrased texts (\autoref{tab:03_prompt_sensitvity}). We find limited differences ($\leq 0.92\%$ for the score and $\leq 0.03$ for the ED).
\begin{table}[!ht]
\small
\begin{center}
\begin{tabular}{lcc}
\toprule
\textbf{Paraphrase Prompt} & 
\makecell{\textbf{PTEB STS}\\\textbf{Score}}  & 
\makecell{\textbf{Edit}\\\textbf{Distance}}  \\
\midrule
Default & 74.92 & 0.81 \\
Variant 1 & 74.24 & 0.82 \\
Variant 2 & 74.82 & 0.82 \\
Variant 3 & 75.16 & 0.84 \\
\bottomrule
\end{tabular}
\caption{PTEB STS scores (English; in \%) and the corresponding edit distances to the original text for the default paraphrase prompt and 3 differently phrased prompt variants; see Appendix \ref{sec:paraphrase_prompts_variations} for the prompts. (Based on all-mpnet-base-v2 embeddings; $n=1$ runs)}
\label{tab:03_prompt_sensitvity}
\end{center}
\end{table}
% 
% 
% ##########################################################################################
% ############################ DISCUSSION ##################################################
% ##########################################################################################
\section{Discussion}
Our method demonstrates that the 3 generative LLMs evaluated in our study are well-suited to rate semantic similarity with average STS scores $>80\%$ (\autoref{tab:01_judge_sts_results}). Gemma-3-27b even outperformed 3 of the 5 embedding models on average on the original STS data (see \autoref{tab:01_judge_sts_results} vs. \autoref{tab:03_sts_embedding_results}).\footnote{With the limitation that Gemma-3-27b was only evaluated on one run and this is, hence, not a like-for-like comparison.} Furthermore, the evaluated LLMs generate paraphrases with high semantic similarity while introducing token-level variation from the original text (\autoref{tab:02_paraphrase_evaluation}). We found that \texttt{gemma3-27b} generated more textually diverse paraphrases at higher throughput: average ED 0.81 vs. 0.55 and 0.63 at a \textasciitilde 5$\times$ to \textasciitilde  20$\times$ speed-up compared to \texttt{gpt-oss-20b} and \texttt{qwen3-32b} respectively. 

Based on our human evaluation for 8 languages, we found that the generated paraphrases are highly similar to the original sentences (mean: 4.05; median: 4.00; see \autoref{tab:human_eval}). However, for Hausa, Swahili, and French, only 70\%, 84\%, and 84\% of sentence pairs met the similarity threshold of 3.00. While this may indicate limited paraphrasing capabilities for low-resource languages like Hausa and Swahili, it is surprising that a high-resource language like French exhibits similar issues.

Across the 25 languages, we observe that all models exhibit lower average scores on PTEB. This drop is most pronounced in English (-2.72pt vs. \mbox{-1.21pt}; \autoref{tab:03_combined_task_results} vs. \autoref{tab:03_multilingual_results}), suggesting that encoders (particularly high-performing ones) may be overfitted to MTEB, relying on shortcuts in the token space rather than semantic understanding. Classification tasks show the largest average performance declines across both English and non-English datasets. We hypothesise that this occurs because classification relies heavily on specific keyword triggers that are disrupted by paraphrasing.

Except for STS22 and RubQ, 18 out of the 20 datasets show lower PTEB scores averaged over all embedding models vs. original scores. For STS22, this can largely be attributed to Embedding-Gemma while the relatively small performance drops for other encoders may be explained by STS22's higher word count (461 vs. 61 for other STS datasets; see \autoref{tab:00_wordcount_sts_results}). Qualitative observations suggest that the longer texts in STS22 are more robust to lexical changes, whereas paraphrases of shorter sentences rely heavily on synonym replacement. In these shorter contexts, individual tokens carry disproportionate weight, potentially distorting the embeddings. For RubQ we believe this is partly due to a higher robustness of the reranking task as also observed for the English datasets (\autoref{tab:03_combined_task_results}, \autoref{tab:03_multilingual_results}). Moreover, RubQ's candidate documents have a substantially higher average word count (63 words vs. 9 for AskUbuntuDuplicateQuestions).

Furthermore, English datasets with initial scores $\geq 80$ (e.g. BIOSSES, SICK-R, STS13/15/17/B, and Banking77) showed relatively large performance gaps (many models -3 to -6pts; also see \autoref{fig:03_mteb_vs_pteb} in the Appendix). This suggests that the initially high scores may depend on shortcuts in token space which are neutralised by paraphrasing. It shows that PTEB's new evaluation method adds a stress test for semantic invariance, mitigating concerns that MTEB may favour lexical shortcuts and overfitted embedding models. However, neither do we claim that any specific model is more overfitted than another one nor do we make general claims on the performance of individual models as the aim of our study is to present a novel evaluation method; not a comprehensive benchmarking study.

Unexpectedly, smaller encoders were not more sensitive to paraphrasing to a certain extend: Our ablation study on the Qwen3 encoder size showed the similar performance for the 4B vs. 8B parameter models (\autoref{tab:03_ablation_emb_models}). For the datasets in our study, this indicates that robust performance under paraphrase stress can be achieved with smaller embedding models. Furthermore, the prompt sensitivity analysis showed that PTEB is robust to prompt changes further mitigating potential concerns that the results might be brittle due to the use of generative LLMs (\autoref{tab:03_prompt_sensitvity}).
%
% ##########################################################################################
% ############################ CONCLUSION ##################################################
% ##########################################################################################
% 
\section{Conclusion}
We introduce PTEB, a sentence embedding benchmark that stochastically paraphrases MTEB data at evaluation time. Using LLM judges and human evaluation, we show that the LLMs in scope yield semantics-preserving yet token-diverse paraphrases. Building on this, we demonstrate that the average performance of the tested embedding models drops across 7 NLP tasks when paraphrased, showing that PTEB effectively uncovers generalisation failures. For non-English languages, we included 6 multilingual datasets covering 25 languages incl. low resource (e.g., African) languages. Ablations over LLM judge and sentence embedding model sizes show runtime–quality trade-offs for the models in scope and that smaller models (12B parameters for LLM judges and 4B for embedding models) perform comparably to larger ones in our study. Our experiments confirm that PTEB yields statistically robust results across multiple runs with different random seeds and even if the paraphrase prompt is varied. In general, PTEB is a novel, more robust evaluation methodology well suited to augment current embedding benchmarks.
%
% ##########################################################################################
% ############################ LIMITATIONS #################################################
% ##########################################################################################
% 
\section{Limitations}

\noindent\textbf{LLM judge and paraphrase model evaluation.} The evaluations of the LLM judge and paraphrase models were limited to a single run. This was sufficient for our goal of demonstrating an end-to-end methodology for PTEB that grounds generative model selection in human gold ratings without human evaluation. For production scenarios, our method can serve as a template and be expanded accordingly. Alternatively, one might pragmatically select a paraphrase model directly.\\ 
 
\noindent\textbf{Human validation.} The human evaluation in our work is limited to 8 languages and 400 sentence pairs (each rated by 2 evaluators). For paraphrasing of English datasets we relied on an automated approach using LLM judges which were only indirectly validated by grounding them in STS gold labels but not validated through human evaluation. \\

\noindent\textbf{PTEB paraphrase model.} Our embedding model evaluation relies exclusively on \texttt{gemma3-27b} paraphrases. While the cross-model-family validation (\autoref{tab:02_paraphrase_cross_eval}) confirmed consistent semantic similarity scores across different LLM judges, future work should examine whether the paraphrasing style itself affects embedding model rankings by running PTEB with different paraphrasers. 

In an initial analysis using the STSB dataset, we compared the edit distance between paraphrases and original texts to the performance drops on PTEB for \texttt{gemma3-27b}, \texttt{gpt-oss-20b}, and \texttt{qwen3-32b} (see \autoref{fig:app_ed_vs_drop} in \autoref{sec:app_add_analyses}). We found that generative models that generate paraphrases with larger edit distances are associated with larger performance drops of embedding models on PTEB. That is, in our initial analysis 4 out of 5 embedding models show the largest performance drops when using \texttt{gemma3-27b} as a paraphrase model (which is the paraphraser generating the largest EDs; see \autoref{tab:02_paraphrase_evaluation}). Similarly, \texttt{qwen3-32b}, which produces the lowest EDs, yields the smallest performance drops for 4 out of 5 embedding models. Despite the need for further analysis, we interpret this as initial evidence that using the ED as a selection criterion is essential, since our goal was to generate paraphrases with substantial token-level changes. Which was exactly one of the reasons to choose \texttt{gemma3-27b} (see \autoref{tab:02_paraphrase_evaluation}) as also illustrated in a non-cherry picked example in \autoref{fig:paraphrase_examples}. 
%
% ##########################################################################################
% ############################ ETHICS ######################################################
% ##########################################################################################
% 
\section{Ethical Considerations}

\noindent\textbf{Model bias.} However, LLMs (and, hence, PTEB) may replicate or amplify biases present in their training data~\citep{Bender2021}. To support ethical scrutiny and reproducibility, we use only open-weight models (i.e., no commercial APIs), enabling researchers to reproduce results, pin versions, and audit behaviour including limited inspection of internals without API accessibility limitations or usage restrictions. While it does not by itself resolve bias, it improves assessability for ethical research. \\

\noindent\textbf{Personal data.} No personal data was collected and all evaluations are based on public datasets. \\

\noindent\textbf{Crowd worker compensation.} We ensured appropriate and equitable compensation by paying all annotators an appropriate and identical rate regardless of their geographic location. At \$0.25 per sentence pair, it corresponds to an estimated hourly wage of \$36. \\

Overall, we did not identify major ethical concerns arising from this work.
%
% ##########################################################################################
% ############################ ACKNOWLEDGEMENTS ############################################
% ##########################################################################################
% 
\section*{Acknowledgments}
We thank the authors of the Massive Text Embedding Benchmark (MTEB)~\cite{Muennighoff2023} and the Massive Multilingual Text Embedding Benchmark (MMTEB)~\citep{MMTEB} for making their code and the included datasets publicly available. While PTEB builds on MTEB and MMTEB, it is an independent extension and is not affiliated with or endorsed by the MTEB or MMTEB authors.

We also thank the anonymous reviewers and human evaluators for their valuable contributions.

We acknowledge the use of AI, such as Anthropic's Claude Code, OpenAI's ChatGPT, and Google's Gemini for assisted coding and writing, e.g., for improving the language of our paper.  

This research was partially supported by the Horizon Europe project GenDAI (Grant Agreement ID: 101182801) and by the ADAPT Research Centre at Munster Technological University. ADAPT is funded by Taighde Éireann – Research Ireland through the Research Centres Programme and co-funded under the European Regional Development Fund (ERDF) via Grant 13/RC/2106\_P2.
\bibliography{custom}        
%
% ##########################################################################################
% ############################ APPENDIX ####################################################
% ##########################################################################################
% 
\appendix
\section{Datasets}
\label{sec:app_datasets}
\autoref{tab:pteb_datasets} shows all PTEB datasets with the respective metrics and the languages included in this study. Further details on all datasets are available in~\citet{Muennighoff2023} and~\citet{MMTEB}, as well as at \url{https://huggingface.co/mteb/datasets}.
\begin{table*}[!htbp]
\small
\centering
\begin{tabularx}{\textwidth}{
  >{\raggedright\arraybackslash}X
  >{\centering\arraybackslash}p{0.25\textwidth}
  >{\centering\arraybackslash}p{0.15\textwidth}
}
\toprule
\textbf{Tasks/Datasets} & \textbf{Languages} & \textbf{Metric} \\
\midrule
% ---- Classification ----
\textbf{Classification} & & Accuracy\\
Banking77 \citep{Banking77Classification} & eng &  \\
AmazonCounterfactuals \citep{AmazonCounterFactuals_Classification} & deu, eng, jpn &  \\
\midrule
% ---- Clustering ----
\textbf{Clustering}  & & V-measure \\
TwentyNewsGroups \citep{TwentyNewsGroupsClustering} & eng &  \\
MasakhaNEWSClusteringS2S \citep{Adelani2024_MasakhaNEWSClusteringS2S} & amh, eng, fra, hau, ibo, lin, lug, orm, pcm, run, sna, som, swa, tir, xho, yor &  \\
\midrule
% ---- Pair Classification ----
\textbf{Pair Classification} & & Average Precision\\
TwitterSemEval2015 \citep{TwitterSemEval2015PairClassification} & eng &  \\
RTE3 \citep{RTE3} & deu, eng, fra, ita &  \\
\midrule
% ---- Reranking ----
\textbf{Reranking} & & MAP\\
AskUbuntuDupQuestions \citep{AskUbuntuDupQPair_Reranking} & eng &  \\
RuBQ Reranking \citep{RubQDataset} & rus &  \\
\midrule
% ---- Retrieval ----
\textbf{Retrieval} (Due to corpus length only queries are paraphrased.) & & nDCG@10\\
ArguAna \citep{ArguAnaRetrieval} & eng & \\
TwitterHjerne \citep{Holm2024_TwitterHernje} & dan &  \\
\midrule
% ---- Summarisation ----
\textbf{Summarisation} & & Spearman corr.\\
SummEval \citep{SummEvalSummarization} & eng &  \\
\midrule
\textbf{Semantic Textual Similarity (STS)} & & Spearman corr. \\
BIOSSES \citep{BIOSSES} & eng &  \\
SICK-R \citep{SICK} & eng &  \\
STS12 \citep{SemEval2012_Task6} & eng &  \\
STS13 \citep{SemEval2013_SharedTask} & eng &  \\
STS14 \citep{SemEval2014_Task10} & eng &  \\
STS15 \citep{SemEval2015_Task2} & eng &  \\
STS17 \citep{SemEval2017_Task1} & eng, ara, deu, fra, ita, nld, spa, tur &  \\
STS22 \citep{SemEval2022_Task8} & eng &  \\
STSBenchmark \citep{SemEval2017_Task1} & eng & \\
\bottomrule
\end{tabularx}
\caption{
Overview of datasets included in PTEB. For each dataset, we report the language(s) and evaluation metric; since we only used the English subset of STS22, other languages are not listed for this dataset.}
\label{tab:pteb_datasets}
\end{table*}
\section{Human Evaluation Details}
\label{sec:app_human_eval}
For each language, we filtered the Amazon Mechanical Turk workers based on their location:
\begin{itemize}
    \item Arabic: Algeria, Egypt, Iraq, Jordan, Lebanon, Morocco, Saudi Arabia, Sudan, Syrian Arab Republic, Tunisia
    \item French: Belgium, Canada, France, Switzerland
    \item German: Austria, Germany, Switzerland
    \item Hausa: Benin, Cameroon, Chad, Ghana, Niger, Nigeria, Togo
    \item Russian: Armenia, Belarus, Brazil, Estonia, Georgia, Germany, Kazakhstan, Kyrgyzstan, Latvia, Lithuania, Russian Federation, Ukraine, United States
    \item Spanish: Argentina, Bolivia, Chile, Colombia, Costa Rica, Dominican Republic, Guatemala, Honduras, Mexico, Panama, Paraguay, Peru, Puerto Rico, Spain, United States
    \item Swahili: Burundi, Congo (DRC), Kenya, Mozambique, Rwanda, Sudan, Tanzania (United Republic of Tanzania), Uganda 
    \item Turkish: Turkey
\end{itemize}

All country names are spelt as on the Amazon Mechanical Turk platform.

The instructions were based on the LLM prompts (see \autoref{sec:app_prompts}.)
Each rating/HIT was paid \$0.25. Based on our assumption that a rating/HIT takes on average 25 seconds, this equals a payment of \$36 per hour. This payment was independent of the worker's location. The time limit for each worker to complete a batch was set to 1 hour. For each language we ran an initial pilot with 20 sentence pairs. In the case of one language, Danish, we were not able to collect any ratings and therefore replaced it with French - while not being a low-resource language, French is a highly relevant for PTEB since 3 of the datasets in this study include French text (see \autoref{tab:pteb_datasets}).  

As a basic validation of the human ratings, we included dummy sentence pairs using an almost identical sentence pair (easy positive) and a completely dissimilar sentence pair (easy negative) in each batch. For each worker, we checked if these were rated correctly within a margin of ±1 before accepting their ratings. Overall, we employed the following validation steps for all ratings based on our 0--5 rating scale:
\begin{enumerate}
    \item Check the ratings of the two dummy sentences.\footnote{Note that not all workers completed the dummy sentence pair ratings.} 
    \item Prompt GPT-5.1 and Claude Sonnet 4.5 with the sentence pairs and human ratings and ask both to flag any ratings which deviate by more than 2.0 from their own similarity estimate.
    \item Identify any sentence pair with a similarity rating of 2.0 and below.
    \item Translate all sentence pairs identified in step 1 to 3 to English using Google Translate and manually inspect the validity of the ratings. 
    \item  Reject workers that fail the dummy sentence pairs within the pre-defined acceptability threshold or repeatedly submitted ratings that deviate by more than 2 from the similarity rating based on the manual inspection and the LLM ratings. Validate the potential candidates to be dropped using the results from step 4.
\end{enumerate}

These validation steps and criteria have been defined a priori to avoid the common mistake of ad-hoc rejection of workers or ratings leading to biased results~\citep{thomson_common_2024}. Generally, we aimed to keep rejections to a minimum. 

Based on these steps, the ratings of exactly one worker for Arabic were rejected since they incorrectly rated the dummy sentence pairs beyond the acceptability threshold. Also, the ratings of the worker in question were consistently more than 2 points below that of other Arabic workers, the two LLMs estimates, and the manual translation-based approach. 

Furthermore, we obtained clarification by two additional workers whose ratings were flagged during the above process while they correctly rated the dummy sentences. Following this, their ratings were ultimately approved.

\section{Implementation Details}
\label{sec:app_implementation_details}
\textbf{Generative Models.}
All generative models were run using Ollama (v0.11.7):
\begin{itemize}
    \item \texttt{qwen3:32b}
    \begin{itemize}
        \item Quantisation: \texttt{Q4\_K\_M}
        \item License: Apache License Version 2.0, January 2004
        \item Link: \url{https://www.ollama.com/library/qwen3:32b}
    \end{itemize}
    
    \item \texttt{gpt-oss:20b}
    \begin{itemize}
        \item Quantisation: \texttt{MXFP4}
        \item License: Apache License Version 2.0, January 2004
        \item Link: \url{https://www.ollama.com/library/gpt-oss:20b}
    \end{itemize}
    
    \item \texttt{gemma3:27b}
    \begin{itemize}
        \item Quantisation: \texttt{Q4\_K\_M}
        \item License: Gemma Terms of Use (last modified 2024-02-21)
        \item Link: \url{https://www.ollama.com/library/gemma3:27b}
    \end{itemize}

    \item \texttt{mistral-small3.2:24b}
    \begin{itemize}
        \item Quantisation: \texttt{Q4\_K\_M}
        \item License: Apache License Version 2.0, January 2004
        \item Link: \url{https://www.ollama.com/library/mistral-small3.2:24b}
    \end{itemize}
\end{itemize}
We used the default hyperparameters and environment variables (\texttt{OLLAMA\_FLASH\_ATTENTION=1}, for example). Further information on the models is available at \url{www.ollama.com/models}.
\\
\\
\textbf{Embedding models.}
All embedding models were ran using the \texttt{sentence-transformers} package~\citep{Reimers2019} (v3.3.1), dtype \texttt{torch.bfloat16} and default \texttt{max\_seq\_len} (ranging from 384 for \texttt{all-mpnet-base-v2} to 32,768 for \texttt{qwen3-embedding-8B}). For instruction-tuned models we use the model-specific prompts if available. Please note that some models like \texttt{qwen3-embedding-8b} might achieve better scores with custom prompts. Further information on the models is available at \url{https://huggingface.co/models}.

For retrieval datasets only the queries were paraphrased due to the length of the corpora. Also, for task-specific hyperparameters we applied the MTEB-defaults (v1.38.33), e.g. \texttt{max\_iter=100} for the Logistic Regression model in the Classification task. For a description of the evaluation protocol for each task the reader may refer to \citet{Muennighoff2023}. MTEB is licensed under the Apache License 2.0.
\\
\\
\textbf{Random Seeds.}
We used \texttt{1337} as a random seed for all experiments. If more than one run was performed, the seed was incremented by one after each run for the paraphrase models only. An exception are the paraphrases of BIOSSES and SICK-R. For these two no paraphrase have been generated with \texttt{seed=1342} but \texttt{seed=1443} instead.
\\
\\
\textbf{Hardware.}
All experiments were conducted on a personal computer with the following specification:
\begin{itemize}
    \item GIGABYTE GeForce RTX 5090 GAMING OC 32G (GPU)  
    \item AMD Ryzen 9 9950X (CPU)
    \item DDR5-6400 64GB (RAM)
\end{itemize}
\Needspace{8\baselineskip}
\section{Prompts}
\label{sec:app_prompts}
The following prompts were used for generative LLMs. Curly brackets indicate placeholders that are filled in with the corresponding variables. These prompts were designed in initial experiments using dummy sentences with the goal to provide the desired output. They are not optimised systematically. The generated paraphrases were selectively reviewed by the authors (in multiple, but not all languages).
\subsection{LLM Judge Prompts}
\label{sec:llm_judge_prompts}
Used in step~1 of our method to rate STS:
\noindent
\begin{tcolorbox}[title=Semantic Similarity Rating Prompt,breakable]
You are an expert evaluator of semantic textual similarity.  
Your task is to rate how similar two sentences are in meaning. \\

Please rate the semantic similarity between the following two sentences  
on a scale from 0.0 to 5.0 using one decimal place:\\

-- 0.0-1.0: Completely different meanings \\  
-- 1.1-2.0: Slightly related but mostly different \\ 
-- 2.1-3.0: Somewhat similar with some shared meaning  \\
-- 3.1-4.0: Very similar with minor differences  \\
-- 4.0-5.0: Essentially the same meaning \\

Only respond with a single decimal number from 0.0 to 5.0 (one decimal place).  
Do not provide explanations or additional text.  \\

Sentence 1: \{sentence1\}  \\
Sentence 2: \{sentence2\}  \\

Similarity rating:
\end{tcolorbox}
\subsection{Paraphrase Evaluation Prompts}
Used in step 2 of our method to rate the paraphrase models:
\label{sec:paraphrase_eval_prompts}
\begin{tcolorbox}[title=Paraphrase Rating Prompt,breakable]
You are an expert evaluator of semantic textual similarity. Your task is to rate how similar two sentences are in meaning.\\

Please rate the semantic similarity between the following two sentences on a scale from 0.0 to 5.0 using one decimal place:\\
-- 0.0-1.0: Completely different meanings\\
-- 1.1-2.0: Slightly related but mostly different\\
-- 2.1-3.0: Somewhat similar with some shared meaning\\
-- 3.1-4.0: Very similar with minor differences\\
-- 4.0-5.0: Essentially the same meaning\\

Original: \{original\}\\
Paraphrase: \{paraphrase\}\\
        
Only respond with a single decimal number from 0.0 to 5.0 (one decimal place). Do not provide explanations or additional text.\\

Similarity rating:
\end{tcolorbox}
\subsection{Paraphrase Prompts}
\label{sec:paraphrase_prompts}
Used in step~2 and step~3 of our method to generate paraphrases:

\begin{tcolorbox}[title=Default,breakable]
Rephrase the following text while keeping its original meaning.  
Only reply with the paraphrased text and only provide a single response —  
no alternatives! Do not include thinking tokens, explanations or notes.
\end{tcolorbox}
Used for multilingual datasets to maintain consistency between input and output language:
\begin{tcolorbox}[title=Multilingual,breakable]
Rephrase the following text while keeping its original meaning.  
Only reply with the paraphrased text and only provide a single response —  
no alternatives! Do not include thinking tokens, explanations or notes. 
Answer in the same language as the input text. Do not translate to English or any other language. 
\end{tcolorbox}

\subsection{Paraphrase Prompt Variatons for Sensitivity Analysis}
\label{sec:paraphrase_prompts_variations}
Used in the prompt sensitivity analysis to test sensitivity to prompt changes. The variants were obtained by promtping GPT-5 to rephrase the default prompt:

\begin{tcolorbox}[title=Variation 1,breakable]
Paraphrase the text below so that the meaning stays the same.  
Return only one rewritten version of the text.  
Do not add explanations, reasoning steps, or multiple options.
\end{tcolorbox}

\begin{tcolorbox}[title=Variation 2,breakable]
Rewrite the given text in different words but preserve its meaning.  
Output exactly one paraphrase.  
Do not include commentary, reasoning, or alternative suggestions.
\end{tcolorbox}

\begin{tcolorbox}[title=Variation 3,breakable]
Provide a single paraphrased version of the following text.  
The meaning must remain unchanged.  
Do not include any notes, thoughts, or more than one response.
\end{tcolorbox}
% % 
% 
\section{Run-Aggregated Statistical Analysis}
\label{sec:app_stat_run_agg_analysis}
\autoref{tab:app_stat_analysis_aggregated} shows the results of the statistical analysis when taking the average over the $n=6$ runs (which reduces the number of observations from $9 \times 6 = 54$ to just $9$) before calculating the HL estimated location shift and applying the Wilcoxon signed-rank test. Due to the smaller number of observations we do not calculate bootstrap CIs.
\begin{table}[!ht]
\small
\centering
\begin{tabularx}{\columnwidth}{l c c}
\toprule
\textbf{all-mpnet-base-v2 vs.}  & \textbf{$\Delta_{HL}$} & \textbf{$p_{Holm}$} \\
\midrule
embeddinggemma-300m & +7.17 & 0.02 \\
\midrule
mxbai-embed-large-v1 & -3.62 & 0.04 \\
\midrule
e5-mistral-7b-instruct & -3.42 & 0.04 \\
\midrule
qwen3-embedding-8b & -4.46 & 0.02 \\
\bottomrule
\end{tabularx}
\caption{Run-aggregated comparison of encoders on PTEB STS vs. \texttt{all-mpnet-base-v2} as baseline. 
$\Delta_{HL}$ is the Hodges–Lehmann estimate of the location shift with its 95\% confidence interval; negative values favour the listed model. P-values are exact Wilcoxon with Holm correction. (9 datasets, average over $n=6$ runs first)}
\label{tab:app_stat_analysis_aggregated}
\end{table}
The run-aggregation follows \citet{Sogaard2014_WhatsPvalueNLP}, who average in their experiments for 2 different NLP tasks over 3 and 10 runs respectively. 

Recommended minimum sample sizes for the bootstrap method vary considerably, from 200~\citep{Sogaard2014_WhatsPvalueNLP} to 20~\citep{Colas2018_HowManyRandomSeeds}, with the requirement that samples be representative of the population~\citep{Colas2018_HowManyRandomSeeds, Dror2018}. In our experiments, the number of observations is $9 \times 6 = 54$  when pooling over datasets and runs (see \autoref{tab:03_stat_analysis}) and $9$ when averaging across runs first (see \autoref{tab:app_stat_analysis_aggregated}). Given these modest number of observations, we use the Wilcoxon signed-rank test for significance testing.

The results show that even in this conservative analysis, the p-values remain below $\alpha = 0.05$ after Holm-correction.
\FloatBarrier
\section{Additional Analyses}
\label{sec:app_add_analyses}
\begin{figure*}[t] 
  \centering
  \includegraphics[width=\linewidth]{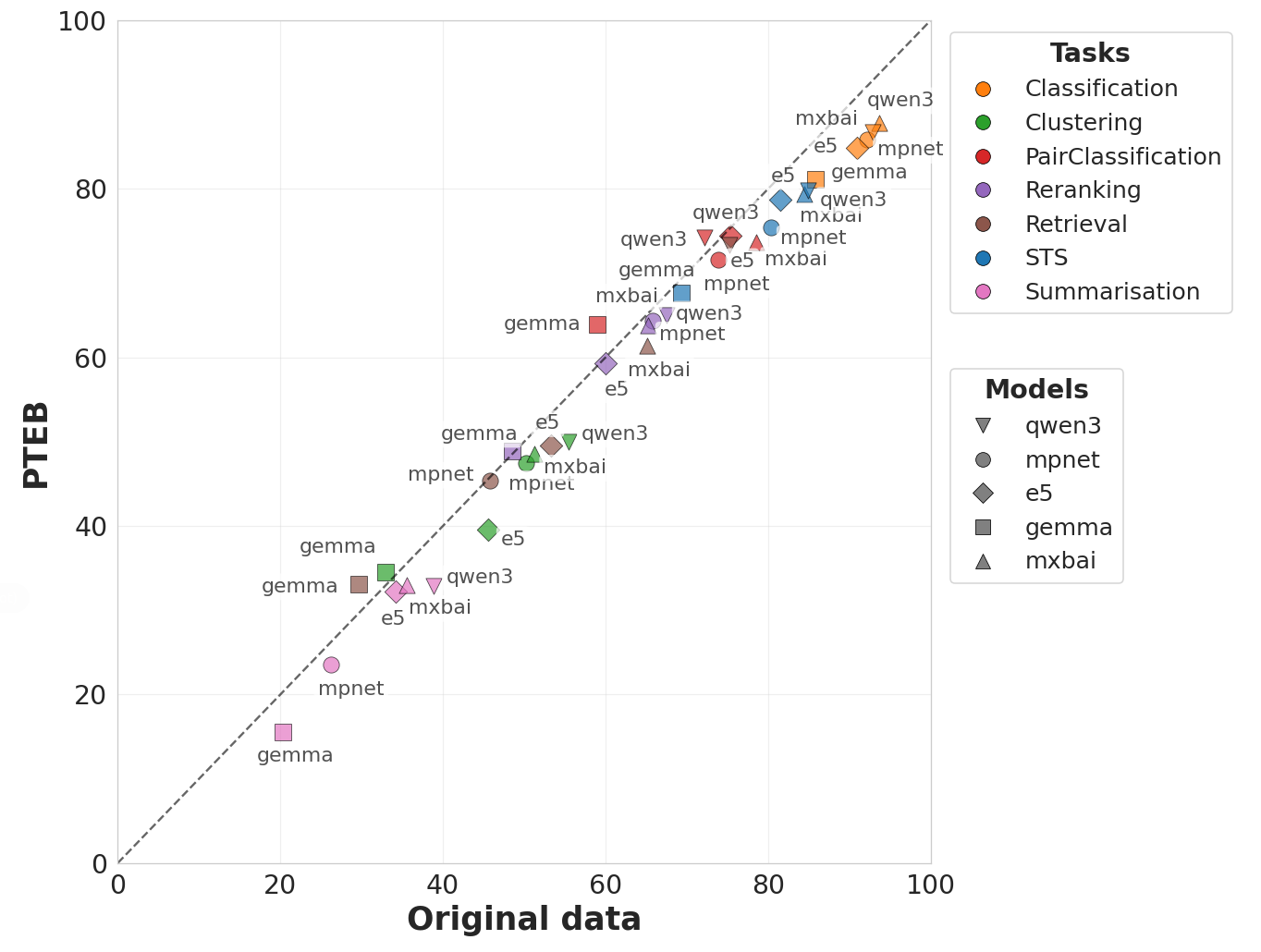}
\captionof{figure}{Scores per task on original MTEB data vs. PTEB (English only, for original MTEB data based on $n=1$ runs, for PTEB based on $n=6$ runs; in \%).}
  \label{fig:03_mteb_vs_pteb}
\end{figure*}

\autoref{fig:03_mteb_vs_pteb} visualises how the performance of embeddings models drops on PTEB vs. original data per for the various task.

\begin{figure*}[t] 
  \centering
  \includegraphics[width=\linewidth]{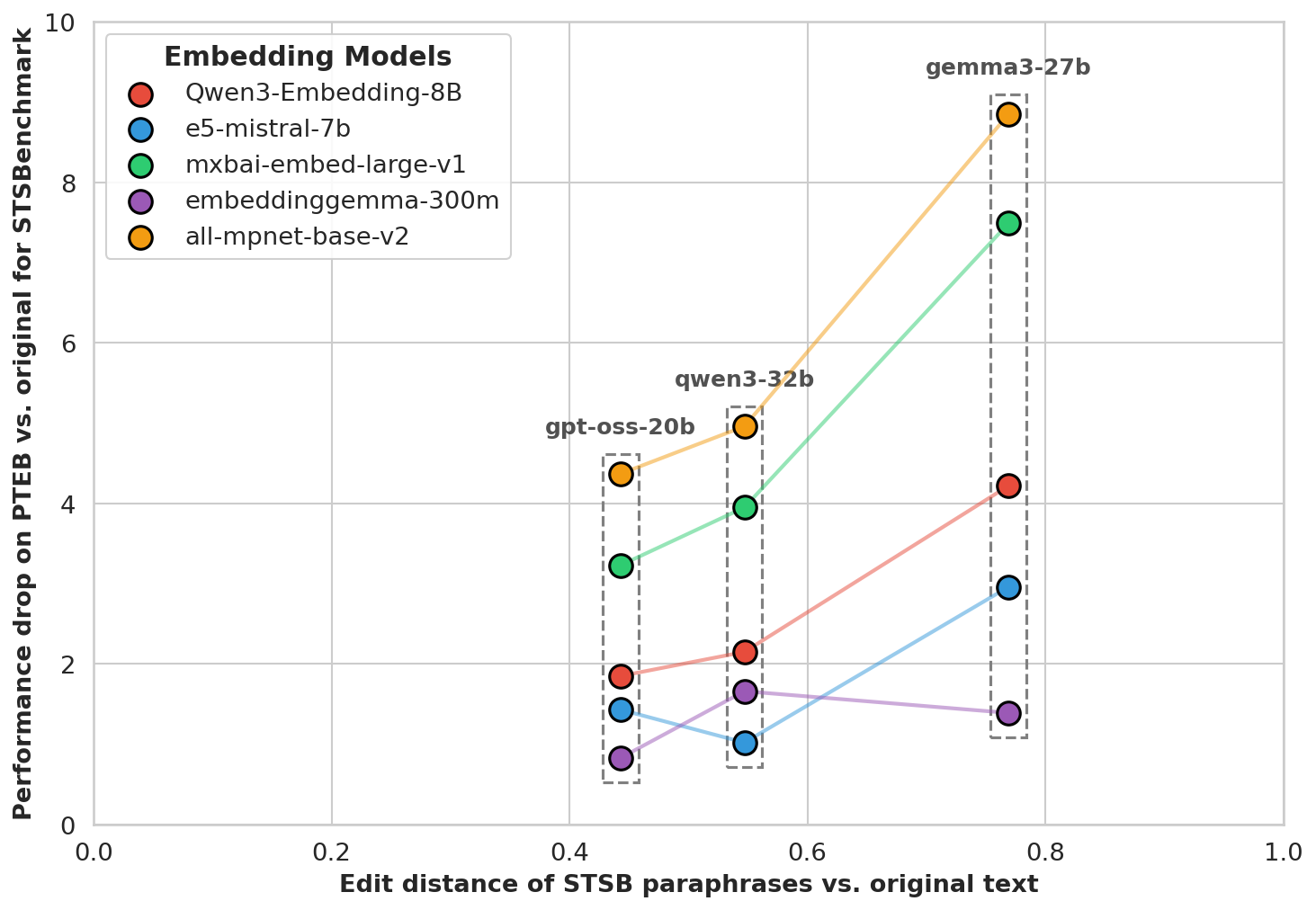}
\captionof{figure}{Edit distance of the paraphrases generated by the different generative paraphrase models (X-axis) vs. the performance drop of the embedding models on PTEB compared to original data for STSBenchmark ($n=1$ runs; in \%).}
  \label{fig:app_ed_vs_drop}
\end{figure*}

\autoref{fig:app_ed_vs_drop} shows how paraphrase models that generate paraphrases with larger edit distance compared to the original data (see \autoref{tab:02_paraphrase_evaluation}) tend to lead to larger performance drops for embedding models on the STSBenchmark dataset.
%
% ##########################################################################################
% ############################ END #########################################################
% ##########################################################################################
% 
\end{document}